\documentclass{ieeeaccess}
\usepackage{cite}
\usepackage{amsmath,amssymb,amsfonts}
\usepackage{algorithmic}
\usepackage{graphicx}
\usepackage{textcomp}

\usepackage{amscd}
\usepackage{wasysym}
\usepackage{mathtools}
\usepackage{stmaryrd}
\usepackage{nameref}
\usepackage[normalem]{ulem}
\usepackage{upgreek}

\usepackage{balance}
\usepackage{hyperref}
\hypersetup{
    colorlinks=true,
    urlcolor=blue,
}

\newcommand\R{\mathbb{R}}
\newcommand\E{\mathbb{E}}
\newcommand\V{\mathbb{V}}

\newcommand{\vect}[1]{\boldsymbol{\mathbf{#1}}}
\newcommand{\func}[1]{\mathrm{#1}}
\newcommand{\matr}[1]{\mathit{#1}}

\DeclareMathOperator*{\argmin}{arg\,min}
\DeclareMathOperator*{\defeq}{\overset{\mathrm{def}}{=}}

\def\BibTeX{{\rm B\kern-.05em{\sc i\kern-.025em b}\kern-.08em
    T\kern-.1667em\lower.7ex\hbox{E}\kern-.125emX}}
\begin{document}
\history{Date of publication xxxx 00, 0000, date of current version December 31, 2020.}
\doi{10.1109/ACCESS.2017.DOI}

\title{ProLab: perceptually uniform projective colour coordinate system}
\author{
\uppercase{Ivan~A.~Konovalenko}\authorrefmark{1,2},
\uppercase{Anna~A.~Smagina}\authorrefmark{1},\\
\uppercase{Dmitry~P.~Nikolaev}\authorrefmark{1,2}, \IEEEmembership{Member, IEEE},
\uppercase{and Petr~P.~Nikolaev}\authorrefmark{1,3}.}
\address[1]{Institute for Information Transmission Problems of the Russian Academy of Sciences, Bolshoy Karetny per. 19, build.1, Moscow, 127051, Russia}
\address[2]{Smart Engines Service, LLC, Prospect 60-Letiya Oktyabrya, 9, Office 605,
Moscow, 117312, Russia}
\address[3]{Moscow Institute of Physics and Technology, 9 Institutskiy per., Dolgoprudny, Moscow Region, 141700, Russia}
\tfootnote{The work was partially financially supported by RFBR in the context of scientific projects N\textsuperscript{\underline{o}}N\textsuperscript{\underline{o}}~17-29-03370 and 19-29-09075.}

\markboth
{Author \headeretal: Preparation of Papers for IEEE TRANSACTIONS and JOURNALS}
{Author \headeretal: Preparation of Papers for IEEE TRANSACTIONS and JOURNALS}

\corresp{Corresponding author: Dmitry~P.~Nikolaev (e-mail: dimonstr@iitp.ru).}

\begin{abstract}
In this work, we propose proLab: a new colour coordinate system derived as a 3D projective transformation of CIE XYZ.
We show that proLab is far ahead of the widely used CIELAB coordinate system (though inferior to the modern CAM16-UCS) according to perceptual uniformity  evaluated by the STRESS metric in reference to the CIEDE2000 colour difference formula.
At the same time, angular errors of chromaticity estimation that are standard for  linear colour spaces can also be used in proLab since projective transformations preserve the linearity of  manifolds.
Unlike in linear spaces, angular errors for different hues are normalized according to human colour discrimination thresholds within proLab. 
We also demonstrate that shot noise in proLab is more homoscedastic than in CAM16-UCS or other standard colour spaces.
This makes proLab a convenient coordinate system in which to perform linear colour analysis.
\end{abstract}



\titlepgskip=-15pt

\maketitle

\section{Introduction}
\label{sec:introduction}

The purpose of this work is to develop a new colour coordinate system for the analysis of colour images of the visible light spectrum.
Most existing algorithms for colour analysis operate with distances in a colour space, and some of them also rely on linear properties of colour distributions.
Meanwhile, the colour space metric, as a rule, is not derived strictly from  physical models, but rather from the properties of human colour perception using psychophysiological experimental data.
The colour coordinate system proposed in this work is based on the colour perception model as well as on  physical models of image formation, so for accurate problem statement we have to provide a detailed introduction.
Sections~\ref{intro_color_spaces} and~\ref{intro_metrics} are dedicated to  colour models in psychophysics; Sections~\ref{intro_cameras} and~\ref{intro_manifolds} to the aspects of colour image formation and processing.
Then, in Section~\ref{intro_statement}, we formulate the problem to be solved and propose the general idea of the solution.
Section~\ref{intro_homography} addresses the most relevant works in the field.
In Section~\ref{intro_properties}, the desired properties of the proposed coordinate system are listed.

\subsection{Colour spaces and colour coordinate systems}\label{intro_color_spaces}
The human perception of colour is defined by the spatial distribution of  retinal irradiance and the internal state of the visual system.
Under photopic conditions, there are three types of active cones, the reactions of which can be considered continuously dependent on the irradiance.
Under the assumption that photoreceptors are negligibly small and uniformly distributed, the response to the irradiance at each point on the retina can be represented as three scalar reactions.
All  other elements of the visual system employ not  arbitrary parameters of the irrandiance, but these three scalar reactions exclusively.
In this model, given a fixed internal state, the colour perception caused by a uniformly illuminated part of the retina depends on a three-dimensional manifold, regardless of the visual system's complexity and its internal state.
We shall call such a manifold a colour space, and its elements we call colours.

To perform colour mapping, various colour coordinate systems are used.
In any of them, the coordinates of the colour vector $\vect{c}$ are related to the spectral irradiance $\func{F}(\lambda)$ through some vector functional $\Psi$, which can be additionally parameterized with  context information and the internal state $\vect{\uptheta}$ of the visual system:
\begin{equation}\label{eq:Psi}
  \vect{c} = \Psi_{\vect{\uptheta}}[\func{F}(\lambda)],
\end{equation}
where $\lambda$ is the light wavelength.
The colour coordinate space is called linear if $\Psi$ is linear for  fixed $\vect{\uptheta}$.

In 1853, Grassmann showed in his experiments~\cite{grassmann1853theorie} that under  colorimetric conditions, a linear 3D colour coordinate system of the human vision system can be constructed, while the internal state of the latter can be neglected due to these conditions.
This reduced the problem of determining the colour space for the human eye to a linear colour coordinate-based construction in the spectral irradiance space $\func{F}(\lambda)$.
In this system, the relationship between the retinal spectral irradiance and the colour coordinates is constructed analytically:
\begin{equation}\label{eq:XYZ}
  \vect{c_x} \defeq \int_{0}^{\infty} \func{F}(\lambda) \vect{X}(\lambda) d\lambda,
\end{equation}
where $\vect{X}(\lambda)$ are the colour matching functions of a standard observer, and $\vect{c_x}$ are the colour coordinates in CIE XYZ.

Thus far, various colour coordinate systems have been proposed for the standard observer.
These systems vary in terms of their suitability for the certain applications~\cite{smith1931cie}.
Some of them imply colorimetric conditions.
Others are related to the various colour perception models, which parametrize $\vect{\uptheta}$ in one way or another.
Assuming that the visual context and the internal state do not affect the spectral sensitivity, the coordinate vectors in any of such coordinate systems can be expressed via CIE XYZ coordinates independently of $\func{F}(\lambda)$:
\begin{equation}\label{eq:short}
  \vect{c_\Phi} = \Psi_{\Phi,\vect{\uptheta}}[\func{F}(\lambda)] = \Phi_{\vect{\uptheta}}(\vect{c_x}),
\end{equation}
where $\Phi_{\vect{\uptheta}}$ is the transformation from CIE XYZ to the given coordinate system under known internal state $\vect{\uptheta}$.

Usually,  colour perception models consider at least the adaptation of the visual system to the dominant illuminance~\cite{fairchild2013color}.
In von Kries' model~\cite{luo2014cie}, this is expressed within the transformation~\eqref{eq:short} as a componentwise division of the input coordinate vector $\vect{c_x}$ by the light source colour coordinate vector $\vect{c_x^{\divideontimes}}(\vect{\uptheta})$:
\begin{equation}\label{eq:kries}
  \Phi_{\vect{\uptheta}}(\vect{c_x}) = \Phi_{\vect{0}}\left(\func{diag}\left(\vect{c_x^{\divideontimes}\left(\vect{\uptheta}\right)}\right)^{-1} \vect{c_x}\right),
\end{equation}
where $\Phi_{\vect{0}}$ is a transformation that is independent of the illumination.

In  systems with the same adaptation model, the coordinate transformation could be performed bypassing CIE XYZ and, obviously, such transformation does not require information on the illumination:
\begin{equation}\label{eq:bypass}
\begin{split}
  &\begin{cases}
    \vect{c_a} = \func{A}_{\vect{\uptheta}}(\vect{c_x}) \defeq \func{A}_{\vect{0}}\left(\func{diag}\left(\vect{c_x^{\divideontimes}\left(\vect{\uptheta}\right)}\right)^{-1} \vect{c_x}\right)\\
    \vect{c_b} = \func{B}_{\vect{\uptheta}}(\vect{c_x}) \defeq \func{B}_{\vect{0}}\left(\func{diag}\left(\vect{c_x^{\divideontimes}\left(\vect{\uptheta}\right)}\right)^{-1} \vect{c_x}\right)
  \end{cases} \\
  &\Longrightarrow \vect{c_b} = \func{B_{\vect{0}}}\left(\func{A_{\vect{0}}}^{-1}(\vect{c_a})\right),
\end{split}
\end{equation}
where $\vect{c_a}$ and $\vect{c_b}$ are the colour coordinate vectors in the two coordinate systems, defined by the transformations $\func{A}_{\vect{\uptheta}}$ and $\func{B}_{\vect{\uptheta}}$, respectively.

\subsection{Evaluation metric and perceptual uniformity}\label{intro_metrics}
Psychophysical experiments not only reveal the spectral basis of the colour space perceived by people, but they also help to determine its metric parameters.
This can be done, for instance, by measuring changes in the thresholds in spectral stimuli for a human eye at different points of the colour space.
The colour coordinate space is called perceptually uniform (hereinafter -- uniform) if the Euclidean distances between colours in it correspond to the differences perceived by a human eye.
Liminal difference vector lengths are uniform across all of the points and in any direction in such a space.
The CIE XYZ linear coordinate system provides a colour space with a natural Euclidean representation, but it is significantly non-uniform in this regard.

There have been many attempts to create a uniform colour coordinate space.
In 1948, Richard Hunter proposed the first uniform space, Hunter Lab~\cite{hunter1948accuracy}.
Later, David MacAdam proposed a space based on a research by Dean Judd~\cite{cie1960international}.
This space was standardized by CIE in 1960 as a uniform chromaticity space (CIE 1960 UCS).
As its name implies, this coordinate system does not include any brightness component.
Soon after that, Gunter Wyszecki proposed a space~\cite{wyszecki1963proposal} based on the latter one, that was adopted as the CIE 1964 (U*, V*, W*) Color Space (or CIEUVW).
It allowed for the calculation of the colour differences even with mismatched brightness.
In 1976, CIELAB space was developed based on Hunter Lab~\cite{mclaren1976development}.
It is still the most used uniform colour coordinate system when it comes to  complex stimuli (images) analysis.

However, the CIELAB space is only approximately uniform, so from the moment of its inception, there have been continuous attempts to create more uniform coordinate systems (e.g.~\cite{kuehni1999towards}).
At the same time, non-Euclidean colour difference formulas were being developed in order to provide more precise correlation with experiment regarding human perception.
The successful outcome of these efforts was the CIEDE2000 formula~\cite{alman2001improvement, luo2001development}, which is still considered the most accurate one available~\cite{wang2012evaluation}.
Nevertheless, for some applications it is preferable to operate in colour coordinate spaces with uniform Euclidean distance (e.g. for the development of  effective search structures).
Therefore, uniform spaces are still being actively researched and developed.
Currently, the CAM16-UCS space~\cite{li2017comprehensive}, developed in 2016, is considered to be the accuracy standard among these.

\subsection{Colour spaces of visible spectrum cameras}\label{intro_cameras}
With certain reservations, all of the aforesaid can  also be attributed to  technical visual systems.
Of course, we are not considering the perception of  technical systems.
In technical vision systems, the term `colour' usually means a three-dimensional vector that is passed for further processing.
Also, the internal state $\vect{\uptheta}$ of such systems, as a rule, could be neglected or considered to be known.
But the most important thing is that the spectral basis of camera space is significantly different from the standard observer ones.
Moreover, colour spaces of cameras made by different manufacturers are usually mismatched.

In order to use the colour coordinates of one colour space within another one, a mapping between these colour spaces must be built.
This concept is invalid  in the general case: an element of any colour space corresponds to an infinite set of metameric spectral irradiance; this set can be mapped into a set of significant volume of the different colour space.

The quality of colour reproduction under different conditions depends on the choice of a particular mapping.
Various mathematical models have been proposed for its construction.
The most commonly used is a linear one~\cite{can2018improving}.
Non-linear colour mappings -- polynomial~\cite{hong2001study} and root-polynomial~\cite{finlayson2015color} -- are also well-known.
Both of these show better accuracy in  experiments.
The latter model is also invariant to changes in brightness, like the linear one.
In~\cite{bianco2013color}, an interesting approach is discussed: the linear mapping parameters are considered to be dependant on the dominant illumination estimation.
The latter is obtained via  analysis of the input image using some algorithm (the dependency of $\vect{\uptheta}$ is established).
Also, in~\cite{kordecki2019practical}  possible refinements of the experimental pipeline are proposed, and in~\cite{vazquez2014perceptual} the choice of loss function minimized for model fitting is discussed.

When dealing with mass-market visible-light cameras, the calibration transformation into standard observer colour space is considered to be known, which allows for the assignment of the human colour space coordinates to the captured colours.
As a rule, the method for the evaluation of the transformation parameters and, sometimes, the model of such a transformation itself are hidden from the user.

\subsection{Importance of linear manifolds in colour image analysis}\label{intro_manifolds}
In technical systems, the algorithms for colour image analysis and processing are based on both human colour perceptual models and physical models of image formation.
The most famous among the latter  is, probably, the dichromatic reflection model proposed by Shafer~\cite{shafer1985using}.
This model assumes that the colour distribution of the uniformly coloured glossy dielectric surface, illuminated by a single source, forms a plane in the linear colour coordinate space.
The assertion of the linear degeneration of the colour distribution was already formulated early, at least by 1975~\cite{nikolaev1975some}, but Shafer's model additionally specified the shape of the colour distribution on the plane.
This model was further developed and generalized, so that the list of conditions under which the colour distributions of uniformly coloured objects form  linear manifolds of various dimensions was expanded~\cite{brill1990image, nikolaev2004linear}.
We shall call all the models of such a family  linear models of colour image formation.

Linear models are used, for instance, in colour based image segmentation~\cite{klinker1988image, cheng2001color, nikolaev2004linear, vinogradova2015image}, as well as in computational colour constancy.
The main problem of the latter is to estimate the colour of the light source of a scene.
One of the ways to solve this is to find the intersection of two dichromatic planes in the linear colour coordinate space.
According to the linear model, the chromaticity of the direction vector of the planes' intersection is the same as the chromaticity of the dominant light source~\cite{lee1986method}.
This technique is used in various colour constancy algorithms~\cite{toro2007multilinear, toro2008dichromatic}.

Note that in earlier works on this topic, only colour distributions forming two-dimensional linear sub-spaces were considered.
However, the algorithms developed later consider a more realistic model which includes diffused light, where the considered manifolds do not pass through the origin of the coordinates~\cite{woo2018improving}.
The generalized versions of such an approach are used in the analysis of scenes with multiple light sources~\cite {nikolaev2004linear, zickler2008color} as well as in  multispectral image analysis~\cite{nikonorov2014spectrum}.

The assumption that the interaction of light with matter is linear leads to a more obvious and fundamental property -- changes in integral illuminance brightness do not affect the chromaticity of image pixels regardless of the number of reflections in the scene, the colouring of the objects, and chromaticity of the light source.
This is what recovery~\cite {finlayson1995color, gijsenij2011computational, hemrit2018rehabilitating} and reproduction angular errors~\cite{finlayson2014reproduction}, widely used in computational colour constancy, are based on.
The first metric considers the angle between the true colour vector of the illuminance and its estimation.
The second  considers the angle between two colour vectors, one of which is the colour vector of a white surface under the given illumination normalized channel-wise by the estimated light source chromaticity, and the other one corresponds to a white surface under an equal-energy illuminant (``true white point”).

Such colour constancy algorithms, based on linear manifold incidence along with angular accuracy errors are applied to linear colour coordinate spaces since in non-linear spaces the relevant geometric properties of the colour distribution are not preserved.

\subsection{The reason behind the development of new colour coordinates}\label{intro_statement}
The structural analysis of linear colour distributions mainly includes two problems.
The first one is to estimate the linear cluster parameters in a colour space using  regression methods, and the key factor here is the tolerance for  colour deviations caused by image noise.
The second problem is the analysis of the mutual positioning of the detected manifolds in a colour space.
The algorithms solving these problems employ the colour differences directly.
So for  applications where these algorithms' behaviour is expected to be in correspondence with  human perception, colour difference metrics should preferably be in correspondence with human ones.

When applying basic statistical methods to the colour distribution of the image, the colour noise is considered to be homoscedastic, i.e. the deviations of colour are additive, well-approximated by a random variable with zero mean, distributed independently from the observed coordinates, and invariant to rotation.
However, even affine transformations of colour coordinates can affect the additive noise anisotropy.
Non-linear transformations can also make the parameters of colour distribution dependant on the observed colour coordinates.
Thus, the correctness of the results obtained via basic statistical methods significantly depends on the coordinate system in which the colour distributions are analysed.

The problem is further complicated by the fact that  image noise is not homoscedastic in the space of linear sensor responses~\cite{bernd2005digital}.
This means that  basic regression methods provide non-optimal estimation of the colour manifolds' positioning even in linear colour spaces.
As a result, both non-linear and linear colour spaces are poorly applicable for  structural analysis: the physical models in  perceptually uniform spaces are over-complicated, and the noise is heteroscedastic~\cite{liang2019spectra}; while in linear spaces the errors are not correlated with  human perceptual differences, and noise homoscedasticity is also not guaranteed for linear spaces.
The same goes for the problem of angular errors: in linear colour spaces,  deviations by the same angle in different directions are not guaranteed to be equally perceived by the human eye, while in perceptually uniform spaces equal chromaticities form a curve, so the concept of angle is poorly applicable here.

So, our goal is to construct a perceptually uniform space of colour coordinates that preserves the linearity of sub-spaces and manifolds.
Moreover, it is preferable that the sensor noise in this space would be as homoscedastic as possible.

\subsection{Homography of 3D colour coordinates}\label{intro_homography}
If the preservation of  manifold linearity required the linearity of the transformation, our goal would be hardly achievable.
First, linear transformations cannot make noise more homoscedastic since the Jacobian matrix of a linear transformation is constant across the space.
(But it is possible to correct the noise anisotropy.)
Also, linear transformations cannot significantly improve perceptual uniformity of the space: colours that are equally spaced in CIE XYZ coordinates are not perceptually uniform on the achromatic axis, and affine transformations preserve the ratio of segment lengths located along any line.

\begin{figure}[t!]
  \centering
  \includegraphics[width=\linewidth]{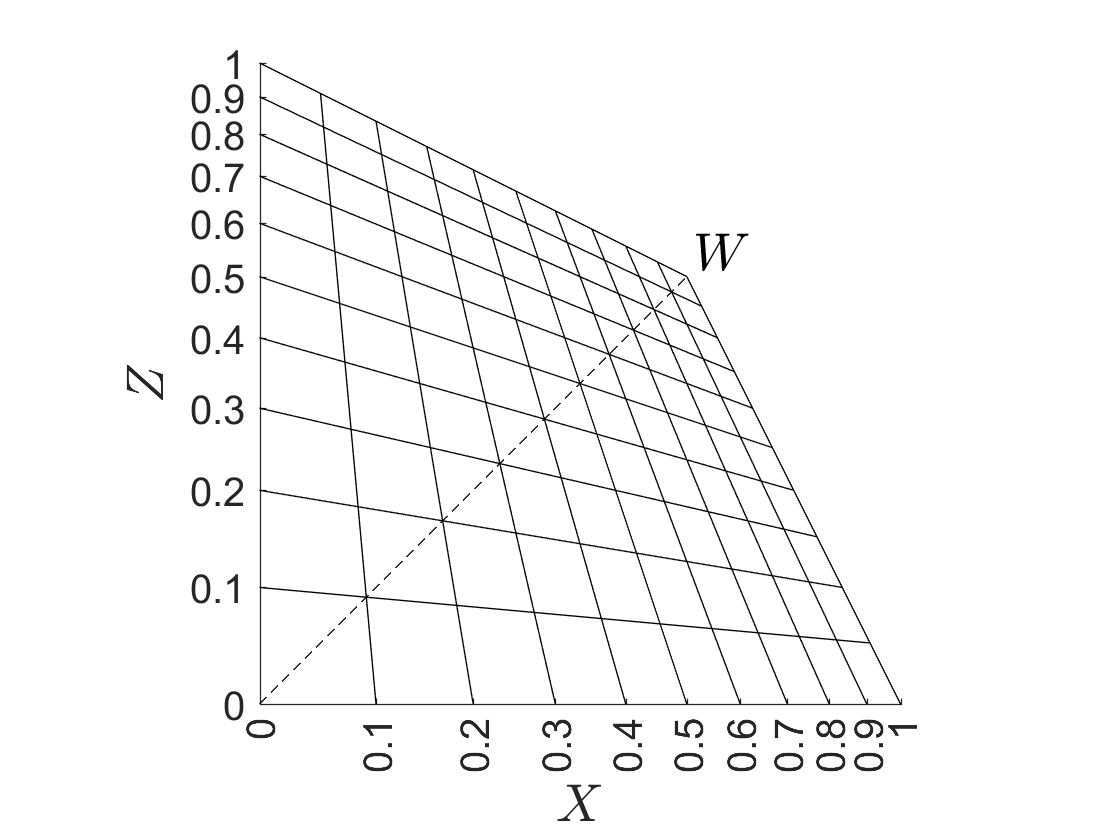}
  \caption{\label{fig:intro}
    XZ plane of the CIE XYZ colour coordinate system in the projective colour coordinate space.
    Point W denotes the projection of the white point.
  }
\end{figure}

But in fact, the class of transformations that preserve the linearity of the manifolds is much wider: it includes any projective transformations.
Unlike affine transformations, projective ones allow for the change of the elements across the space in different ways.
Fig.~\ref{fig:intro} illustrates a projective transformation compressing the space in the vicinity of one point and stretching the space in another while keeping all of the lines straight.
This hints that a solution with  acceptable uniformity is possible, i.e. a colour coordinate system with the desired properties can be developed.

Projective transformations of colour coordinates are already widely used in various applications~\cite{finlayson2019color}.
However, projective transformations are usually applied only to the chromaticity plane.
Apparently, MacAdam~\cite{macadam1937projective} was the first researcher who proposed projective transformations for the construction of a uniform colour space, but in his pioneering work as well as in later papers, only two dimensional transformations are considered.

In 2003, the first work introducing the transformation of the entire colour space was published~\cite{wallace2003color}.
It employed  3D projective transformation for colour gamut matching for various projector devices.
Later, the same approach was applied to photo-realistic colour transfer between images~\cite{gong20193d}.
Both cases employ mutual calibration of two images rather than transforming into some reference space with a different metric.
The latter method was used in work  which shows that fixed 3D dimensional projective colour coordinate transformation improves the results of colour-based image segmentation~\cite{smagina2019linear}.

\subsection{Proposed colour coordinate system}\label{intro_properties}
In this work, we propose proLab: a uniform colour coordinate system for the standard observer which is based on a three-dimensional projective transformation of CIE XYZ coordinates.
We demonstrate the following advantages of the new coordinate system:
\begin{itemize}
  \item ProLab is far ahead of the commonly used uniform system CIELAB according to perceptual uniformity (although inferior to CAM16-UCS).
  \item The image shot noise  is more homoscedastic in proLab than in other uniform spaces.
  \item Among the uniform systems, ProLab is the only one which preserves the linearity of  colour manifolds; this property allows for the employment of  angular accuracy of colour reproduction and for  linear colour analysis in accordance with a human colour difference metric.
  \item The transformation from CIE XYZ to proLab has an elegant analytical expression, and it is computationally efficient compared with CAM16-UCS and the even more primitive CIELAB.
\end{itemize}

We presented the idea of constructing the proLab coordinate system for the first time at 25th Symposium of the International Society for Color Vision in 2019~\cite{konovalenko2019prolab}, but the construction methodology and a numerical study of proLab properties in detail are discussed for the first time in the current work .
Moreover, the model parameters have been estimated more accurately compared to the one presented at the Symposium.

\subsection{Outline of the main results: paper structure}
The main part of the article is organized as follows.
Four sections are focused on the construction of the proLab coordinate system.
In Section~\ref{sect:proLab_general}, we introduce the necessary notation, construct the basic model of the transformation from CIE XYZ to proLab, and discuss how to determine the transformation parameters that do not affect the metric.
In Section~\ref{sect:proLab_restrictions}, a priori restrictions on proLab metric parameters are given.
In Section~\ref{sect:uniformity}, we introduce a function quantifying the perceptual uniformity of  colour coordinates.
In Section~\ref{sect:methodology}, we provide the optimal proLab parameters along with the step-by-step methodology to obtain them.

Then, we study the properties of the resulting colour coordinate system.
In Section~\ref{sect:homoscedasticity}, we propose a function to quantify the deviation of the colour coordinate noise parameters from  homoscedasticity.
In Section~\ref{sect:noise}, the noise model for the colour sensor is constructed, and  noise parameters estimated on a raw image from an open dataset are provided.
In Section~\ref{sect:comparison}, we provide a numerical comparison of proLab properties with the existing coordinate systems in terms of  uniformity and noise homoscedasticity.
In the Discussion, we review some qualitative properties of the proposed coordinate space and also suggest possible optimizations of proLab parameters.
In the Conclusion, the main results of this work are summarized.

\section{Basics of proLab colour model}\label{sect:proLab_general}
Let us now construct a colour coordinate space of the standard observer, so that colour manifolds linear in CIE XYZ will remain linear in the constructed space as well.
In addition, we require the Euclidean distance in this space to approximate the colour differences determined in a certain way.

Let us denote the CIE XYZ colour coordinate space as $C_x$, the CIELAB coordinate space as $C_l$, and the constructed space as $C_p$.
By $\func{L}$ we denote the transformation from $C_x$ to $C_l$, and by $\func{P}$ from $C_x$ to $C_p$.
As follows from the requirement of the manifolds' linearity preservation, $\func{P}$ is a three-dimensional projective transformation.
Let us parametrize any projective transformation in standard matrix notation denoted by italics (e.g. matrix $P\defeq(p_{ij})\in\R^{4\times4}$ corresponds to the transformation $\func{P}$).

Let $\func{T_h}$ and $\func{T_c}$ be functions for transformation between Cartesian and homogeneous coordinates:
\begin{equation}\begin{split}
  &\func{T_h}(\vect{c}) \defeq \begin{bmatrix}\vect{c} \\ 1\end{bmatrix}, \quad \vect{c} \in \R^3, \\
  &\func{T_c}(\vect{h}) \defeq \begin{bmatrix}I_3 & \vect{0}\end{bmatrix} \frac{\vect{h}}{h_4}, \quad \vect{h} \in \R^4, \quad h_4 \ne 0,
\end{split}\end{equation}
where $I_3$ is the $3 \times 3$ identity matrix.
Then
\begin{equation}\label{eq:homography_form_1}
  \func{P}(\vect{c_x}) \defeq \func{T_c}\big(P\func{T_h}(\vect{c_x})\big), \quad \vect{c_x} \in C_x, \quad \func{P}(\vect{c_x}) \in C_p.
\end{equation}

Any requirements on metrics over $C_p$ define $\func{P}$ only up to a similarity in $C_p$.
Indeed, applying a rigid transformation to $C_p$ does not change distances, while isotropic scaling is equivalent to a change in the distance measuring units.
To make the solution unique, we need to introduce some additional restrictions on $\func{P}$.

Let us build the coordinate system that could replace CIELAB in the simplest possible way.
First of all, we require distances in $C_p$ to model the dominant light source adaptation in the same way as in CIELAB.
For this purpose, let us consider the simplified von Kries model~\cite{luo2014cie}, performing the component-wise division of the input vector coordinates $\vect{c_x}$ by the light source colour coordinates $\vect{c_x^{\divideontimes}} \in C_x$.
Let us denote the adaptation transformation as $\func{N}$.
This transformation is projective, and its corresponding transformation matrix can be defined as:
\begin{equation}\label{eq:N_form}
  N \defeq \func{diag}\left(\func{T_h}\left(\vect{c_x^{\divideontimes}}\right)\right)^{-1}.
\end{equation}
Now $\func{P}$ can be decomposed into the adaptation transformation $\func{N}$ and the transformation $\func{Q}$ independent of the dominant light source:
\begin{equation}\label{eq:PQ_adaptation}
  \func{P} = \func{Q} \circ \func{N}, \quad P = Q N, \quad Q \defeq (q_{ij}) \in \R^{4\times4},
\end{equation}
where $\circ$ is the function composition operator.
In addition, we require the black point $\vect{0}$ to be preserved by $\func{P}$:
\begin{equation}\label{eq:black_zero}
  \func{P}(\vect{0}) = \vect{0}.
\end{equation}
Then $Q$ can be written as follows:
\begin{equation}\label{eq:Q_decomposition}
  \matr{Q}(\vect{\upvarphi}, \rho, \vect{\upmu}) = \matr{R_1}(\varphi_1) \matr{R_2}(\varphi_2) \matr{R_3} (\varphi_3) \matr{Z}(\rho) \matr{M}(\vect{\upmu}),
\end{equation}
where $\matr{R_i}(\varphi)$ is a matrix for rotation by an angle $\varphi$ about an axis $i$, $Z(\rho)$ is the isotropic scaling matrix with a scaling factor $\rho > 0$, and $M(\vect{\upmu})$ is a special matrix which defines $C_p$'s metric properties:
\begin{equation}\label{eq:M_form}
  M (\vect{\upmu}) \defeq \begin{bmatrix}
    \mu_1 & \mu_2 & \mu_3 & 0 \\
    0     & \mu_4 & \mu_5 & 0 \\
    0     & 0     & 1     & 0 \\
    \mu_6 & \mu_7 & \mu_8 & 1
  \end{bmatrix}, \quad |M| > 0,
\end{equation}
where $\vect{\upmu}\in\R^8$ is a vector of metric parameters.

Decomposition~\eqref{eq:Q_decomposition} allows us to determine the metric parameters $\vect{\upmu}$ separately from the similarity parameters $\vect{\upvarphi}$ and $\rho$.
Here, we do not consider mirroring of the coordinate system, as the simultaneous fulfilment of conditions $|M| > 0$ and $\rho > 0$ itself keeps the colour hues' traversal order around the achromatic axis the same as in CIE XYZ and CIELAB.
Also note that the matrices $R_i(\varphi)$ and $Z(\rho)$ are defined up to multiplication by a nonzero scalar.
In order to define these matrices, we require the bottom right element to be equal to one.
Hence,
\begin{equation}\label{eq:p=q=1}
  p_{44} = q_{44} = 1.
\end{equation}

Consider the metric parameters $\vect{\upmu}$ to be known.
Now, let us fix the rest of the parameters.
First, let us set the direction and overall scale of the lightness axis as in CIELAB:
\begin{equation}\label{eq:L_axis_orientation}
  \func{P}\left(\vect{c_x^{\divideontimes}}\right) = \begin{bmatrix} 100 & 0 & 0 \end{bmatrix}^T.
\end{equation}

\begin{figure}[t!]
  \centering
  \includegraphics[width=\linewidth]{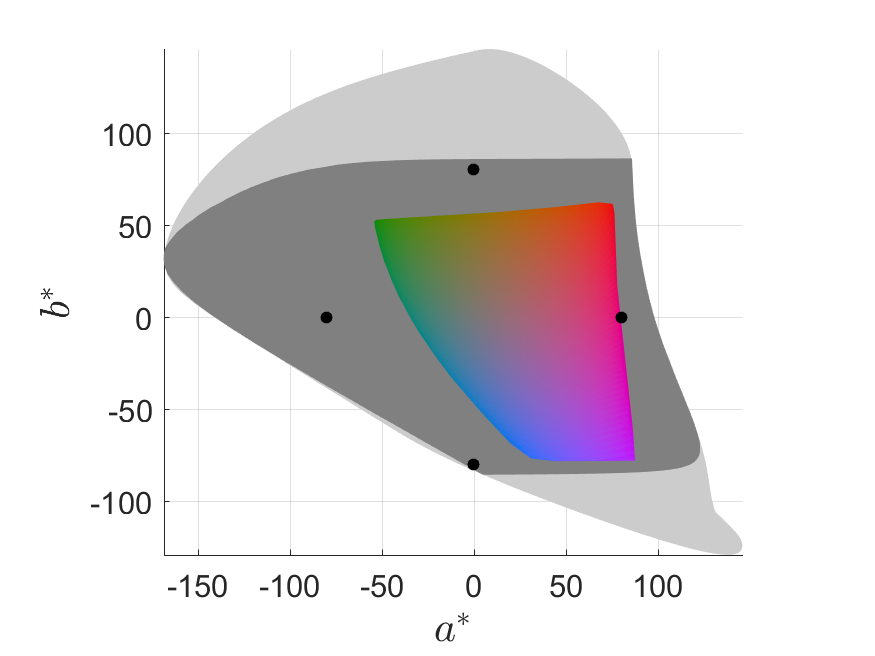}
  \caption{\label{fig:four_points}
    Projection of the four selected hue orientation points onto the plane $(a^*, b^*)$.
    The light grey region corresponds to the D65 gamut projection, the dark grey corresponds to its section by the plane $L^*=50$ containing the selected points, and the same section of the sRGB gamut is shown as the coloured region.
  }
\end{figure}

This condition defines parameters $\varphi_2$, $\varphi_3$, and $\rho$ unambiguously, while the rotation angle $\varphi_1$ around the lightness axis is still undefined.
In order to fix this parameter as well, let us require the hues of the saturated colours to be arranged approximately similar to CIELAB.
Specifically, let us choose four equally saturated CIELAB points $C_l^{key} \subset C_l$ within the D65 gamut, such that for each of them the lightness is equal to the half of the maximum, and one colour coordinate is equal to zero (see Fig.~\ref{fig:four_points}):
\begin{equation}
  C_l^{key} \defeq \left\{\begin{bmatrix} 50 \\ -80 \\ 0 \end{bmatrix}, \begin{bmatrix} 50 \\ 80 \\ 0 \end{bmatrix}, \begin{bmatrix} 50 \\ 0 \\ -80 \end{bmatrix}, \begin{bmatrix} 50 \\ 0 \\ 80 \end{bmatrix}\right\}.
\end{equation}
Then we choose $\varphi_1$ so that the coordinates of the selected points in $C_p$ differ from those in $C_l$ as little as possible:
\begin{equation}\label{eq:ab_axes_orientation}
  \varphi_1^{opt} = \argmin_{\varphi_1} \sum_{\vect{c_l} \in C_l^{key}} \left\| \vect{c_l} - \func{T_c}\Big(P(\vect{\upvarphi}, \rho,\vect{\upmu})\func{T_h}\big(\func{L^{-1}}(\vect{c_l})\big)\Big) \right\|_2^2.
\end{equation}
The $\varphi_1^{opt}$ can be found analytically by finding optimal rotation~\cite{besl1992method}.

Summarizing the above, we fully determine proLab as the transformation $\func{P}$ constructed via the following steps:
\begin{itemize}
  \item The optimal metric parameters $\vect{\upmu}$ of the matrix $M$~\eqref{eq:M_form} are calculated according to the metric requirements.
  \item The scale and the lightness axis orientations are set to be the same as in CIELAB~\eqref{eq:L_axis_orientation}.
  \item The colour axes orientation is determined according to~\eqref{eq:ab_axes_orientation}.
  \item Normalization~\eqref{eq:PQ_adaptation} is taken into account.
\end{itemize}

\section{Additional restrictions on proLab parameters}\label{sect:proLab_restrictions}
Now let us define a subspace of the metric parameters, $\vect{\upmu}$, over which the proLab colour model can be interpreted meaningfully.
First of all, we should note that not every projective transformation maps the original gamut into a bounded region.
Let us require this natural property.

In the space $C_x$, let a plane given by the equation
\begin{equation}\label{eq:horizon}
  \begin{bmatrix} p_{41} & p_{42} & p_{43} \end{bmatrix} \vect{c_x} + p_{44} = 0,
\end{equation}
be called a horizon of the space $C_p$.
On the horizon, the denominator of the rational transformation $\func{P}$ vanishes.
The gamut image in $C_p$ is a bounded region if and only if the preimage of this gamut in $C_x$ does not intersect with the horizon.

Now let us formulate a simple sufficient condition under which the gamut is bounded in $C_p$.
This condition does not require knowledge of the gamut shape.
Note that the gamut of any light source in $C_x$ always lays within an orthotropic rectangular box, the main diagonal of which connects vertices $\vect{0}$ and $\vect{c_x^{\divideontimes}}$.
All eight vertices of such a box can be listed as follows:
\begin{equation}\label{eq:box}
  \func{diag}\left(\vect{c_x^{\divideontimes}}\right) \vect{b}, \quad \vect{b} \in \{0, 1\}^3.
\end{equation}
Taking into account the normalization~\eqref{eq:p=q=1}, the condition on the gamut to be bounded in $C_p$ can be written as the condition on all these vertices to be on the same side with regards to the horizon~\eqref{eq:horizon}:
\begin{equation}\label{eq:constraint_on_P:horizon}
  \begin{bmatrix} p_{41} & p_{42} & p_{43} \end{bmatrix} \func{diag}\left(\vect{c_x^{\divideontimes}}\right) \vect{b} + 1 \ge 0, \quad \vect{b} \in \{0, 1\}^3.
\end{equation}
Now let us determine $\vect{\upmu}$ under which this condition is fulfilled.
From~\eqref{eq:N_form},~\eqref{eq:PQ_adaptation} and~\eqref{eq:Q_decomposition}, it follows that
\begin{equation}
\begin{split}
  &\begin{bmatrix} \mu_6 & \mu_7 & \mu_8 \end{bmatrix} = \begin{bmatrix} q_{41} & q_{42} & q_{43} \end{bmatrix}=\\
  &= \begin{bmatrix} p_{41} & p_{42} & p_{43} \end{bmatrix} \func{diag}\left(\vect{c_x^{\divideontimes}}\right),
\end{split}
\end{equation}
so the condition~\eqref{eq:constraint_on_P:horizon} can be rewritten as
\begin{equation}\label{eq:constraint:horizon}
  \begin{bmatrix} \mu_6 & \mu_7 & \mu_8 \end{bmatrix} \vect{b} + 1 \ge 0, \quad \vect{b} \in \{0, 1\}^3.
\end{equation}
Note that for $\vect{b} = \vect{0}$ this condition is fulfilled for any $\vect{\upmu}$.

Now let us introduce one more restriction.
As with the coordinate $L^*$ of the CIELAB space, we would like the first coordinate of proLab to represent the lightness.
Therefore, we require the following:
\begin{equation}\label{eq:brightness_sanity}
\begin{split}
  &\vect{0} \le \vect{c_0}, \vect{c_1} \le \vect{c_x^{\divideontimes}}, \ \vect{c_1} - \vect{c_0} \in \R_{\ge 0}^3 \\
  &\Longrightarrow \left(\func{P}(\vect{c_1}) - \func{P}(\vect{c_0})\right)^T \vect{\hat{e}_L} \ge 0,
\end{split}
\end{equation}
where $\vect{\hat{e}_L} \defeq \begin{bmatrix} 1 & 0 & 0 \end{bmatrix}^T$ is the lightness unit vector (here and hereafter, we denote inequality for a vector as a system of inequalities restricting each of the coordinates).
This restriction means that within the bounding box of the gamut, increase of any coordinate in CIE XYZ should not lead to a decrease in the lightness coordinate in proLab.

Let us simplify the requirement~\eqref{eq:brightness_sanity}.
To do this, we consider a set of planes with equal lightness (the first coordinate) in proLab.
In $C_p$, they can be expressed as
\begin{equation}
  L^+ = L^+_s, \quad L^+ \defeq \vect{\hat{e}_L}^T\vect{c_p}.
\end{equation}
Since $\func{P}$ is projective, the preimage of the set of these planes forms a pencil of planes in $C_x$.
At $0 \le L^+_s \le 100$, the preimages of planes in $C_x$ cross the rectangular box~\eqref{eq:box}.
Since we have already required the condition~\eqref{eq:constraint_on_P:horizon}, we can consider the angle of the preimage rotation around the pencil axis to be a monotonic and continuous function of $L^+_s$ within the given range.
Then condition~\eqref{eq:brightness_sanity} is equivalent to the requirement that the coordinates of the normals should be of one sign.
Since the sign of a normal's coordinates cannot be changed twice, we can instead require all coordinates of normals to the extreme planes (with equations $L^+ = 0$ and $L^+ = 100$) to be non-negative.

Let us define the restrictions on $\vect{\upmu}$ under which this requirement is satisfied.
To do that, we introduce another coordinate space $C_b$, which is constructed by the projective transformation $\func{B}$ of the space $C_x$ with a matrix $\matr{B} \defeq \matr{M}\matr{N}$:
\begin{equation}\label{eq:C_b}
  \func{B}(\vect{c_x}) \defeq \func{T_c}(\matr{M}\matr{N}\func{T_h}(\vect{c_x})), \quad \vect{c_x} \in C_x, \quad \func{B}(\vect{c_x}) \in C_b.
\end{equation}
From~\eqref{eq:PQ_adaptation} and~\eqref{eq:Q_decomposition} it follows that
\begin{equation}
  \matr{P} = \matr{R_1}\matr{R_2}\matr{R_3}\matr{Z}\matr{B},
\end{equation}
i.e. the space $C_b$ is related to the space $C_p$ through a similarity.

Let us denote the parameters of the plane given by the equation $\vect{l_\Phi} \func{T_h}(\vect{c_\Phi}) = 0$ in an arbitrary colour coordinate space $C_\Phi$ ($\vect{c_\Phi} \in C_\Phi$) as the vector $\vect{l_\Phi}$.
According to the definition~\eqref{eq:C_b}, the following relationship holds between parameters $\vect{l_x}$ of the planes in the $C_x$ space and parameters $\vect{l_b}$ of the images of these planes in the $C_b$ space:
\begin{equation}\label{eq:l_x}
  \vect{l_x} = \vect{l_b} M N.
\end{equation}
So, the restriction on the parameters of the line in the space $C_b$ leads to the requirement of non-negative normal coordinates of this line image in the space $C_x$, and this requirement can be written as follows:
\begin{equation}
  \vect{l_b} MN \ge \begin{bmatrix} 0 & 0 & 0 & -\infty \end{bmatrix}.
\end{equation}
The matrix $N$ is diagonal with positive elements, so the inequality can be be simplified as follows:
\begin{equation}\label{eq:non-zero}
  \vect{l_b} M \ge \begin{bmatrix} 0 & 0 & 0 & -\infty \end{bmatrix}.
\end{equation}

Now, let us apply this restriction to the planes $L^+ = 0$ and $L^+ = 100$ in $C_p$.
For this, let us consider the white point image in the space $C_b$.
Let us denote it as $\vect{c_b^{\divideontimes}}$:
\begin{equation}
  \vect{c_b^{\divideontimes}} \defeq \func{B}\left(\vect{c_x^{\divideontimes}}\right).
\end{equation}
Since the planes of pencil $L^+ = L^+_s$ in the space $C_p$ are orthogonal to the white point vector $\vect{c_p^{\divideontimes}}$ in this space, their preimages are also orthogonal in the space $C_b$.
This means that the parameters of the planes of such a pencil in $C_b$ are expressed as
\begin{equation}
  \begin{bmatrix} \vect{c_b^{\divideontimes}}^T & l_4\left(L^+_s\right) \end{bmatrix},
\end{equation}
where $l_4\left(L^+_s\right)$ denotes the dependency of the fourth coordinate of the plane parameters' vector on $L^+_s$.

The preimage of plane $L^+=0$ passes through $\vect{0}$, while the preimage of plane $L^+=100$ passes through the white point $\vect{c_b^{\divideontimes}}$ in the space $C_b$.
Hence, the parameters of the corresponding preimages are equal to $\begin{bmatrix} \vect{c_b^{\divideontimes}}^T & 0 \end{bmatrix}$ and $\begin{bmatrix}\vect{c_b^{\divideontimes}}^T & - \vect{c_b^{\divideontimes}}^T \vect{c_b^{\divideontimes}} \end{bmatrix}$, respectively.

Let us substitute these parameters into the condition~\eqref{eq:non-zero} and expand $\matr{M}$ in terms of the definition~\eqref{eq:M_form}:
\begin{equation}\label{eq:non-zero-2}
  \begin{bmatrix}
    \vect{c_b^{\divideontimes}}^T & 0 \\
    \vect{c_b^{\divideontimes}}^T & -\vect{c_b^{\divideontimes}}^T \vect{c_b^{\divideontimes}}
  \end{bmatrix}
  \begin{bmatrix}
    \mu_{1} & \mu_{2} & \mu_{3} \\
    0       & \mu_{4} & \mu_{5} \\
    0       & 0       & 1       \\
    \mu_{6} & \mu_{7} & \mu_{8}
  \end{bmatrix} \ge 0_{2,3}.
\end{equation}

From the definition~\eqref{eq:N_form}, it follows that  $N\func{T_h}\left(\vect{c_x^{\divideontimes}}\right) = \vect{1}$, so
\begin{equation}\label{eq:white_in_m_from_C_x}
  \vect{c_b^{\divideontimes}} = \func{T_c}(M N\func{T_h}\left(\vect{c_x^{\divideontimes}}\right)) = \func{T_c}(M \vect{1}).
\end{equation}
Taking into account~\eqref{eq:M_form}, we get
\begin{equation}
\begin{split}
  &\vect{c_b^{\divideontimes}} = \vect{m}/m, \quad
  \vect{m} = \begin{bmatrix} \mu_1 + \mu_2 + \mu_3 \\ \mu_4 + \mu_5 \\ 1 \\ \end{bmatrix}, \\
  &m = \mu_6 + \mu_7 + \mu_8 + 1.
\end{split}
\end{equation}
Due to the condition~\eqref{eq:constraint:horizon} $m \ge 0$, the inequality~\eqref{eq:non-zero-2} is equivalent to
\begin{equation}\label{eq:constraint:intensity}
  \begin{bmatrix} m \vect{m}^T & 0 \\ m \vect{m}^T & - \vect{m}^T \vect{m} \end{bmatrix} \begin{bmatrix}
    \mu_{1} & \mu_{2} & \mu_{3} \\
    0 & \mu_{4} & \mu_{5} \\
    0 & 0 & 1 \\
    \mu_{6} & \mu_{7} & \mu_{8}
  \end{bmatrix} \ge 0_{2,3}.
\end{equation}

Finally, taking into account the $7$ restrictions on $\vect{\upmu}$ given in~\eqref{eq:constraint_on_P:horizon} and the $6$ restrictions given in~\eqref{eq:constraint:intensity}, we obtain $13$ additional restrictions on the proLab parameters in a form
\begin{equation}
  f_i(\vect{\upmu}) \ge 0,
\end{equation}
where $f_i(\vect{\upmu})$ are polynomials of third or smaller degree.

\section{Perceptual uniformity criteria}\label{sect:uniformity}
In the region bounded by the above inequalities, let us find a vector $\vect{\upmu}$ which maximizes perceptual uniformity.
The perceptual uniformity of a colour coordinate space implies the accuracy of the perceptual colour differences approximation by Euclidean distances in this space.
In order to quantify non-uniformity, the STRESS (STandardized REsidual Sum of Squares) criterion is usually employed: the higher the STRESS value, the worse the uniformity~\cite{kruskal1964multidimensional, garcia2007measurement, wang2012evaluation, pan2018comparative}.

Let $\vect{a}$ denote the vector of colour differences estimated in one approximation, and let $\vect{b}$ denote the vector of the same colour differences estimated in some other approximation, such that $\|\vect{a}\| \ne 0$ and $\|\vect{b}\| \ne 0$.
The STRESS criterion for these two vectors is defined as follows:
\begin{equation}\label{eq:STRESS}
  \func{STRESS}\left(\vect{a}, \vect{b}\right) \defeq \frac{\|k \vect{a} - \vect{b}\|_2}{ \|\vect{b}\|_2}, \quad k = \frac{(\vect{a}, \vect{b})}{\|\vect{a}\|_2^2},
\end{equation}
and it is easy to see that the STRESS criterion is equal to the absolute value of the sine of the angle between $\vect{a}$ and $\vect{b}$:
\begin{equation}
  \func{STRESS}\left(\vect{a}, \vect{b}\right) = \sqrt{1 - \frac{(\vect{a}, \vect{b})^2}{\|\vect{a}\|_2^2 \ \|\vect{b}\|_2^2}} = \left|\sin\overset{\scalebox{2}[0.5]{$\wedge$}}{\vect{a}\vect{b}}\right|.
\end{equation}

This criterion is symmetric and invariant to scaling by either of the two compared estimations:
\begin{equation}
  \func{STRESS}\left(k\vect{a}, \vect{b}\right) = \func{STRESS}\left(\vect{b}, \vect{a}\right), \quad k\ne0.
\end{equation}
Hence, fixing the colour coordinate scale as  given by the condition~\eqref{eq:L_axis_orientation} does not affect  correspondence evaluation via the STRESS criterion.

This criterion is also invariant to equal permutations of colour difference vector components:
\begin{equation}
  \func{STRESS}\left(M_\pi\vect{a}, M_\pi\vect{b}\right) = \func{STRESS}\left(\vect{a}, \vect{b}\right),
\end{equation}
where $M_\pi$ is an arbitrary permutation matrix.
Thus, STRESS can be defined on a multiset of ordered difference pairs.
Let us consider $\omega = \{(a_i, b_i) \mid 1 \le i \le n\}$, $\vect{a}, \vect{b} \in \R^n$ as a finite sample of ordered pairs of real numbers.
The STRESS criterion for this sample is defined in an obvious way:
\begin{equation}
  \func{STRESS}\left(\omega\right) \defeq \func{STRESS}\left(\vect{a}, \vect{b}\right).
\end{equation}

The STRESS value can be significantly dependent on the distribution of measured samples.
The values measured in perceptual experiments (as in~\cite{wang2012evaluation}), or obtained using colour difference formulas (as in~\cite{thomsen2000euclidean, urban2007embedding}) are used as a reference.
The disadvantage of the first approach is the fixed and limited number of samples (3657 colour pairs among all datasets as of 2001~\cite{luo2001development}), so here  the question arises about the space coverage and the sample distribution balance.
The disadvantage of the second approach is the additional approximation error.
In this work we use the second approach, employing the CIEDE2000 formula~\cite{alman2001improvement, luo2001development} as the reference.

Following~\cite{urban2007embedding}, let us construct the pairs from the colour vectors uniformly distributed within the light source gamut in the CIELAB space.
However, unlike~\cite{urban2007embedding}, we do not place an upper bound on the colour difference in each pair as we do not want to be limited to the analysis of small differences only.

Let us denote the D65 gamut as $G \subset C_l$, and the uniform sample from this gamut as $G_n$ ($G_n \subset G, |G_n| = n$).
We shall similarly denote the uniform sample of the gamut's colour pairs as $G^2_n$ ($G^2_n \subset G^2 \subset C_l^2, |G^2_n| = n$).
Finally, we denote the reference CIEDE2000 colour difference of the colour pair $p \in C_l^2$ as $\func{{\Delta E}^*_{00}}(p)$; as $\Phi$ -- the transformation into the considered space $C_\Phi$ from  $C_x$, and as $\func{\Phi_L}$ -- the same from $C_l$ ($\func{\Phi_L} = \Phi \circ \func{L^{-1}}$), where $\circ$ is the composition of transformations.
Then the criterion of non-uniformity of the colour coordinate space $C_\Phi$ over the sample $G^2_n$ can be written as follows:
\begin{equation}\begin{split}\label{eq:U_crit}
  &\func{U}[\Phi, G^2_n] \defeq \func{STRESS}(\omega), \\
  &\omega = \left\{\left(\left\|\func{\Phi_L}(\vect{c_a}) - \func{\Phi_L}(\vect{c_b})\right\|_2, \func{{\Delta E}^*_{00}}(p)\right) \mid p = \left(\vect{c_a},\vect{c_b}\right) \in G^2_n\right\}.
\end{split}\end{equation}

\section{Optimal proLab parameters}\label{sect:methodology}
To obtain proLab parameters according to all the aforesaid, we performed the following steps:

\begin{enumerate}
  \item We computed the grid of points on the surface of the D65 gamut $G$, using the method described by V.~Maksimov in~\cite{maksimov1984transformation}.
  Two-dimensional triangulation was constructed for these points.
  This allowed $G$ to be approximated  by a polyhedron with $\sim 20 \ 000$ faces.
  Since $G$ is a convex set, in further analysis we used a system of linear inequalities to check whether the points were inside of the gamut $G$.
  Each of the inequalities verifies the position of points relative to one of the faces.

  \item The sample $G_{2n_1}$ consisting of $2n_1$ independent and identically distributed CIELAB colours that belonged to $G$ gamut was generated, where $n_1 = 10 \ 000$.
  By dividing the sample $G_{2n_1}$ into $n_1$ pairs, a sample of $G^2_{n_1}$ pairs was formed.

  \item To determine the metric parameters $\vect{\upmu}^{opt}$, we solved the optimization problem with penalty functions~\cite{back1997handbook}:
  \begin{equation}\label{eq:general_proLab_problem}
  \begin{split}
    &\vect{\upmu}^{opt} = \argmin_{\vect{\upmu} \in \R^8} \func{U}[C_m(\vect{\upmu}), G^2_{n_1}]+\\
    &+ \Sigma \sum_{i=1}^{14} \max(0, -f_i(\vect{\upmu}))^2,
  \end{split}
  \end{equation}
  where $\Sigma$ is a parameter of the penalty function method, $\vect{f}$ is a vector of  functions corresponding to the conditions given above: $f_1 = |M| = \mu_1 \mu_4$ -- to the condition~\eqref{eq:M_form}, $\{f_2, ..., f_8\}$ -- to $7$ non-trivial linear conditions~\eqref{eq:constraint:horizon}, and $\{f_9, ..., f_{14}\}$ -- to $6$ cubic conditions~\eqref{eq:constraint:intensity}.
  To calculate the criterion $\func{U}$, we obtained values $\func{{\Delta E}^*_{00}}$ according to procedures described in~\cite{sharma2005ciede2000}.
  The problem~\eqref{eq:general_proLab_problem} was solved numerically via multistart sequential quadratic programming~\cite{marti2018multi, nocedal2006numerical}.
  As a result, we obtained the following matrix of metric parameters~\eqref{eq:M_form}:
  \begin{equation}
    M = \begin{bmatrix*}[r]
      2.1591 & -1.7823 & -0.0713 & 0 \\
      0      &  2.0866 &  0.2103 & 0 \\
      0      &  0      &  1      & 0 \\
      0.7554 &  3.8666 &  1.6739 & 1
    \end{bmatrix*}.
  \end{equation}

  \item To fully determine matrix $Q$~\eqref{eq:Q_decomposition}, we found parameters $\vect{\upvarphi}$ and $\rho$ analytically (see section~\ref{sect:proLab_general}).
  As a result, the following matrix $Q$ was formed:
  \begin{equation}\label{eq:Q_final}
    Q = \begin{bmatrix*}[r]
      75.5362  &  486.661  &  167.387 & 0 \\
      617.7141 & -595.4477 & -22.2664 & 0 \\
      48.3433  &  194.9377 & -243.281 & 0 \\
      0.7554   &  3.8666   &  1.6739  & 1
    \end{bmatrix*}.
  \end{equation}
\end{enumerate}

So, for a D65 light source with coordinates $\vect{c_x^{\divideontimes}} = \begin{bmatrix} 0.9505 & 1 & 1.0888 \end{bmatrix}^T$~\cite{ohta2006cie}, we can obtain the following proLab parameters:
\begin{equation}\label{eq:P_final}
  P = \begin{bmatrix*}[r]
    79.4725  &  486.6610 &  153.7311 & 0 \\
    649.9038 & -595.4477 & -20.4498  & 0 \\
    50.8625  &  194.9377 & -223.4334 & 0 \\
    0.7947   &  3.8666   &  1.5373   & 1
  \end{bmatrix*}.
\end{equation}

\section{Criterion of noise heteroscedasticity}\label{sect:homoscedasticity}
As we mentioned in the introduction, colour values captured by a camera are always noisy.
As a rule, statistical methods for colour distribution analysis consider the image noise as homoscedastic.
To validate this, let us construct a criterion to estimate the heteroscedasticity of colour vector noise in the space of colour coordinates.

Let us consider the colour coordinate space $C_\Phi$ with known transformation $\Phi: C_x \to C_\Phi$.
We assume that at each point $\vect{c_x} \in C_x$ the noise is approximately additive with zero mean and a known covariance matrix $\matr{\Sigma_x}(\vect{c_x})$.
Then, covariance of noise in $C_\Phi$ space could be roughly expressed as follows:
\begin{equation}\label{eq:any_covariance}
  \matr{\Sigma_\Phi}(\vect{c_\Phi}) = J_\Phi(\vect{c_x}) \  \matr{\Sigma_x}\left(\vect{c_x}\right) \ J_\Phi^T(\vect{c_x}), \quad \vect{c_x} = \func{\Phi^{-1}}\left(\vect{c_\Phi}\right),
\end{equation}
where $J_\Phi$ is the Jacobian matrix of the transformation $\Phi$.

We consider the noise to be homoscedastic if all three eigenvalues of its covariance matrix are equal throughout the gamut.
Then, on the colour sample $G_n$, we can estimate the hardware noise heteroscedasticity in space $C_\Phi$ as follows:
\begin{equation}\begin{split}\label{eq:H_crit}
  &\func{H}[\Phi, G_n] \defeq \func{STRESS}(\omega), \\
  &\omega = \left\{\left(\lambda_i^{1/2}\left[\matr{\Sigma_\Phi}(\func{\Phi_L}(\vect{c_l}))\right], 1\right) \mid \vect{c_l} \in G_n, 1 \le i \le 3\right\},
\end{split}\end{equation} 
where $\lambda_i[\matr{A}]$ is the $i$-th eigenvalue of a matrix $\matr{A}$.

\section{Noise parameters in sensor colour space and other spaces}\label{sect:noise}
Let us construct a noise model for the original sensor colour space.
A quite simple model of output values for a single-channel image, which nevertheless agrees well with  experiments,   was proposed by J\"{a}hne in~\cite{bernd2005digital}:
\begin{equation}\label{eq:Yane_model}
  s = g \, n + \varepsilon, \quad n \sim \func{Pois}(s_0), \quad \E(\varepsilon) = 0,
\end{equation}
where $s$ is a random sensor response, $g$ is a gain coefficient, $n$ is a random number of registered electrons, $s_0$ is an expected sensor response value at $g = 1$, and $\varepsilon$ is  additive noise independent of the sensor irradiance.

\begin{figure}[t!]\begin{center}
  \includegraphics[width=0.7\linewidth]{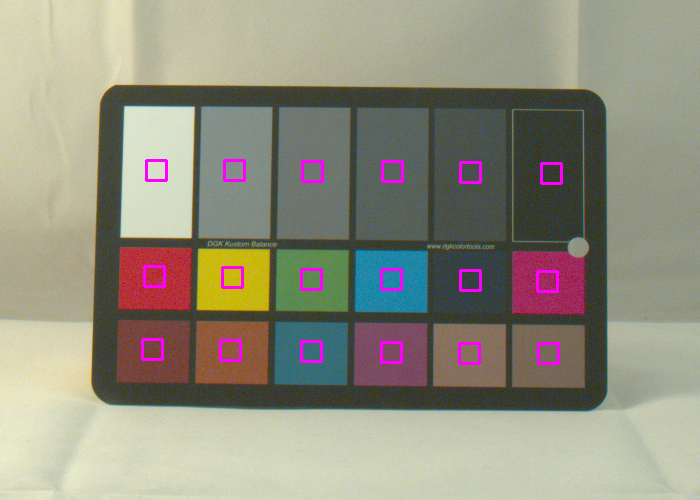}
  \caption{\label{fig:colorchecker_sampling}
    Colorchecker captured using Canon 5D Mark III camera.
    The areas used for the estimation of sensor response mean and variance are marked with violet rectangles.
  }
\end{center}\end{figure}

According to~\eqref{eq:Yane_model}, the relationship between the mean and variance of output values is linear:
\begin{equation}\label{eq:Yane_noise_model}
  \V(s) = g \, \E(s) + \V(\varepsilon).
\end{equation}
To verify this model, let us take the MLSDCR (Multiple Light Source Dataset for Colour Research) dataset~\cite{smagina2020multiple}, which was captured using a Canon 5D Mark III camera.
Among various scenes, MLSDCR contains raw images of the colorchecker (see Fig.~\ref{fig:colorchecker_sampling}); parameters of the calibration transformation from the camera colour space into the standard observer colour space (in  sRGB coordinates) are also provided.

\begin{figure}[t!]
  \centering
  \includegraphics[width=\linewidth]{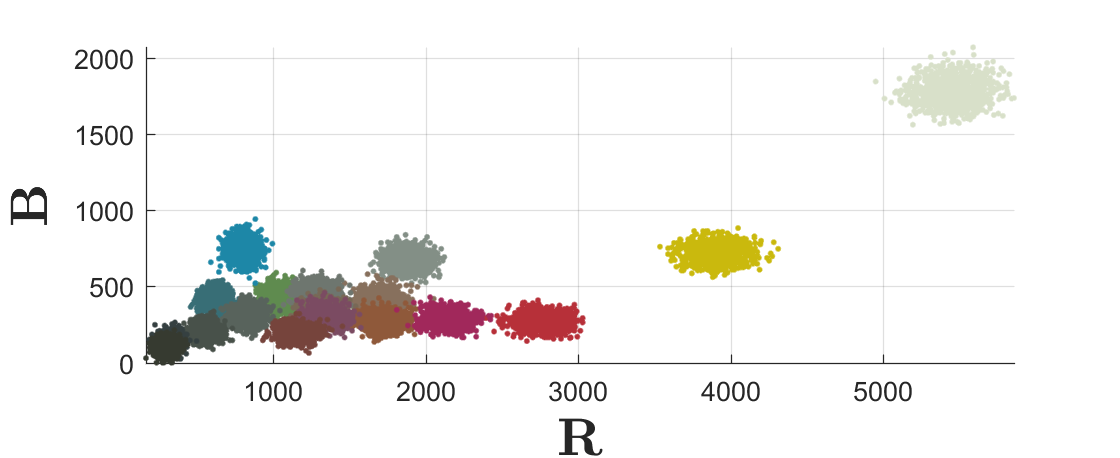}
  \caption{\label{fig:heteroscedasticity_example_all}
    Sensor responses for each patch of the colour chart.
  }
\end{figure}

Let us estimate the noise parameters for each patch (uniformly coloured area) of the colour chart.
The measurement accuracy of colour calibration experiments may be limited by irradiance non-uniformity~\cite{kordecki2019practical}.
Thus, we need to track irradiance uniformity -- but only inside each patch, not between them.
To reduce the impact of irradiance non-uniformity, we take a small central area of $42\times42$ Bayer mosaic pixels (see Fig.~\ref{fig:colorchecker_sampling}) for each patch of the colour chart.
We also take into account that different mosaic elements can have different noise parameters.
In the Canon 5D Mark III camera, a standard (RGGB) Bayer mosaic is used, so for each of the $18$ patches we shall form a sample $S_i$ ($1 \le i \le 72$) of sensor responses for uniform irradiance.
Joint histograms of colour coordinates through the sample elements demonstrate significant noise heteroscedasticity (see Fig.~\ref{fig:heteroscedasticity_example_all}).

\begin{figure}[t!]
  \centering
  \includegraphics[width=\linewidth]{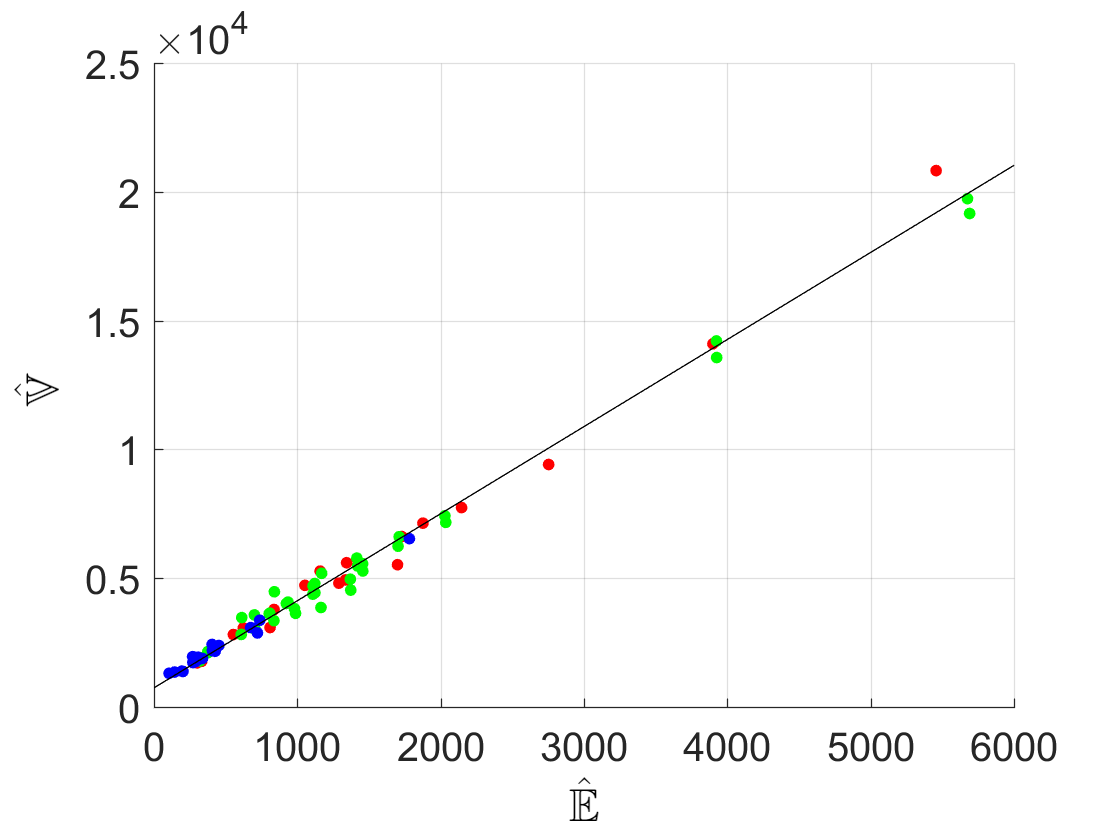}
  \caption{\label{fig:Yane_model}
    Linear dependency of the sensor response variance on its average sample value.
    Colours of points correspond to the conventional names of sensor channels.
  }
\end{figure}

Let us estimate the mean and variance of the sensor responses for each element of $S_i$:
\begin{equation}
  \hat{\E}_i = \overline{S_i}, \quad \hat{\V}_i = \overline{(S_i-\overline{S_i})^2}, \quad 1 \le i \le 72.
\end{equation}
Using principal component analysis, we estimate parameters of the model given in~\eqref{eq:Yane_noise_model}:
\begin{equation}
  \hat{g} = 3.38, \quad \hat{\V}(\varepsilon) = 744.
\end{equation}
The relationship between the sample estimations $\hat{\E}(s)$ and $\hat{\V}(s)$ can be approximated with good reliability by linear dependency (see Fig.~\ref{fig:Yane_model}).
Thus, we estimate the variance in Bayer mosaic value captured using the Canon 5D Mark III camera as follows:
\begin{equation}\label{eq:Yane_noise_model_special}
  \hat{\V}(s) = 3.38 \, s + 744.
\end{equation}

Let us consider the transformation into the space $C_x$.
We use the transformation described in~\cite{smagina2020multiple}, from where the experimental data was obtained.
In this work, the sensor colour coordinate space is obtained via the simplest de-bayering algorithm, which employs an averaging of two mosaic G-elements and aggregates the average with a single R- and a single B-element.
Let us call such a space  `deviceRGB' and denote it as $C_d$.
Taking into account the noise model described above~\eqref{eq:Yane_noise_model_special}, the covariance matrix of the noise in $C_d$ could be written as
\begin{equation}\label{eq:sensor_covar}
\begin{split}
  &\matr{\Sigma_d}(\vect{c_d}) = \func{diag}\left(\begin{bmatrix}1 & 1/2 & 1\end{bmatrix}\right)\left(3.38 \, \func{diag}(\vect{c_d}) + 744 \, I_3\right),\\ &\vect{c_d} \in C_d.
\end{split}
\end{equation}

The transformation matrix from deviceRGB to the standard observer linRGB is described in~\cite{smagina2020multiple}:
\begin{equation}
  D_1 = \displaystyle \frac{0.03}{2^{16}} \begin{bmatrix}
     41.93 & -2.08  & -37.24 \\
    -14.32 &  39.13 &  10.79 \\
    -0.02  & -35.39 &  185.52
  \end{bmatrix}.
\end{equation}
Considering that the transformation from linRGB to CIE XYZ is linear~\cite{stokes1996standard} and given by the matrix
\begin{equation}
  D_2 \defeq \begin{bmatrix}
    0.4125 & 0.3576 & 0.1804 \\
    0.2127 & 0.7152 & 0.0722 \\
    0.0193 & 0.1192 & 0.9503
  \end{bmatrix},
\end{equation}
we obtain the following transformation matrix from $C_d$ to $C_x$:
\begin{equation}\label{eq:device_transform}
  D^{-1} = D_2 \, D_1 = 10^{-6} \begin{bmatrix}
     5.5711 &  3.0892  & 10.0585 \\
    -0.6066 &  11.4383 & 6.0363  \\
    -0.4189 & -13.2786 & 80.9631
  \end{bmatrix}.
\end{equation}
The sensor noise covariance matrix in CIE XYZ space is equal to
\begin{equation}\label{eq:XYZ_sensor_noise}
    \matr{\Sigma_x}(\vect{c_x}) = D^{-1} \ \matr{\Sigma_d}(D \vect{c_x}) \ D^{-T}, \quad \vect{c_x} \in C_x.
\end{equation}

By substituting the parameter values from~\eqref{eq:sensor_covar} and~\eqref{eq:device_transform} in~\eqref{eq:XYZ_sensor_noise}, we obtain the numerical colour noise model for our camera in space $C_x$.

\section{Comparing the performance of proLab with other perceptually uniform colour spaces}\label{sect:comparison}
We estimate the perceptual non-uniformity further according to the criterion~\eqref{eq:U_crit}: $\func{U_T}[\Phi] \defeq \func{U}[\Phi, G^2_{n_2}]$, where $\Phi$ is the transformation from CIE XYZ by which the system is defined.
In order to do this, we form an independent test sample $G^2_{n_2}$ of $n_2 = 100 \ 000$ pairs, according to the method described in section~\ref{sect:methodology}.

Let us also construct a test $G_{n_2}$ of individual colours to estimate noise heteroscedasticity.
In doing so, we require all of the colour vectors not only to belong to the D65 source gamut, but also to be reproducible by the camera -- i.e. we require each component of the colour vectors to be non-negative in deviceRGB space:
\begin{equation}
  \vect{c_l} \in G_{n_2} \Longrightarrow D\func{L^{-1}}(\vect{c_l}) \ge 0.
\end{equation}
We perform uniform sampling $G_{n_2}$ from a reproducible subarea of the gamut and estimate the heteroscedasticity according to the criterion~\eqref{eq:H_crit}: $\func{H_T}[\Phi] \defeq \func{H}[\Phi, G_{n_2}]$.
To calculate $\func{H_T}[\Phi]$, we need to use the noise covariance matrix $\matr{\Sigma_\Phi}(\vect{c_\Phi})$, which we obtain using  approximation~\eqref{eq:any_covariance} along with the colour noise model~\eqref{eq:XYZ_sensor_noise} in $C_x$ space.

\begin{table}
\caption{Performance of the colour coordinate systems.
Bold underlined font indicates the best criteria results; bold only -- second best.}
\label{tabular:comparison}
\setlength{\tabcolsep}{3pt}
\begin{tabular}{|p{50pt}|p{50pt}|p{50pt}|p{50pt}|}
\hline
$C_\Phi$ space& \textbf{Collineation} & $\func{U_T[\Phi]}$ & $\func{H_T[\Phi]}$ \\
\hline
    LMS & \underline{\textbf{Yes}} & $0.475$ & $0.720$ \\
    deviceRGB & \underline{\textbf{Yes}} & $0.474$ & $\underline{\vect{0.470}}$ \\
    CIE XYZ & \underline{\textbf{Yes}} & $0.479$ & $0.722$ \\
    CIE xyY & 
    \textbf{Central pencil} & $0.296$ & $0.822$ \\
    linRGB & \underline{\textbf{Yes}} & $0.381$ & $0.607$ \\
    sRGB & No & $0.316$ & $0.830$ \\
    CIELAB & No & $0.259$ & $0.848$ \\
    CAM16-UCS & No & $\underline{\vect{0.177}}$ & $0.696$ \\
    proLab & \underline{\textbf{Yes}} & $\vect{0.209}$ & $\vect{0.565}$ \\
\hline
\end{tabular}
\label{tab1}
\end{table}

The proposed colour coordinate space was further compared with the following ones:
\begin{itemize}
  \item CIE XYZ~\cite{smith1931cie} -- basic colour coordinate system of the standard observer;
  \item CIE xyY~\cite{smith1931cie} -- system with  distinct chromaticity coordinates;
  \item LMS~\cite{fairchild2013color} -- coordinate system that models linearized human cone responses;
  \item sRGB~\cite{stokes1996standard} -- colour coordinates used to represent colours on displays and printers (most photos and videos are coded with these coordinates);
  \item linRGB~\cite{stokes1996standard} -- intermediate (without gamma correction) representation of reproduced colours linearly related to CIE XYZ;
  \item CIELAB~\cite{mclaren1976development} -- widely spread perceptually uniform colour coordinates;
  \item CAM16-UCS~\cite{li2017comprehensive} -- the most perceptually uniform coordinate system at the time of writing.
\end{itemize}

We compare these colour spaces and linear colour coordinates of the camera sensor with proLab via the criteria $\func{U_T}$ and $\func{H_T}$.
The results of quantitative comparison using $\func{U_T}$ and $\func{H_T}$ are demonstrated in Table~\ref{tabular:comparison}.
We also specify whether each of the colour coordinate systems preserves linearity of the colour manifolds or not.
ProLab preserves lines by its construction as well as linear coordinate systems; only CIE xyY has a nontrivial classification by collineation, since it keeps  lines passing through $\vect{0}$ as lines.
The rest of the colour coordinate systems do not keep even the central pencil linear.
As to  perceptual uniformity, proLab is inferior to CAM16-UCS -- the modern and currently the most accurate space -- but it is significantly superior to the common CIELAB uniform space.
Also, our experiments show that proLab is inferior in terms of noise homoscedasticity only to the deviceRGB space, the properties of which vary significantly from camera to camera.

\begin{figure*}[t!]
  \centering
  
  \minipage{0.33\linewidth}\center{\includegraphics[width=\linewidth]{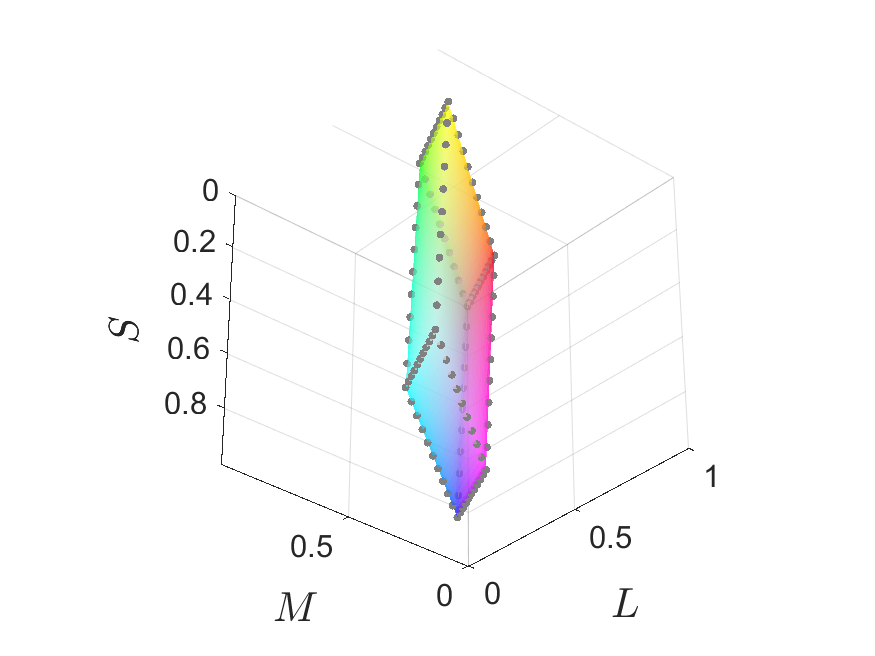}}\endminipage
  \minipage{0.33\linewidth}\center{\includegraphics[width=\linewidth]{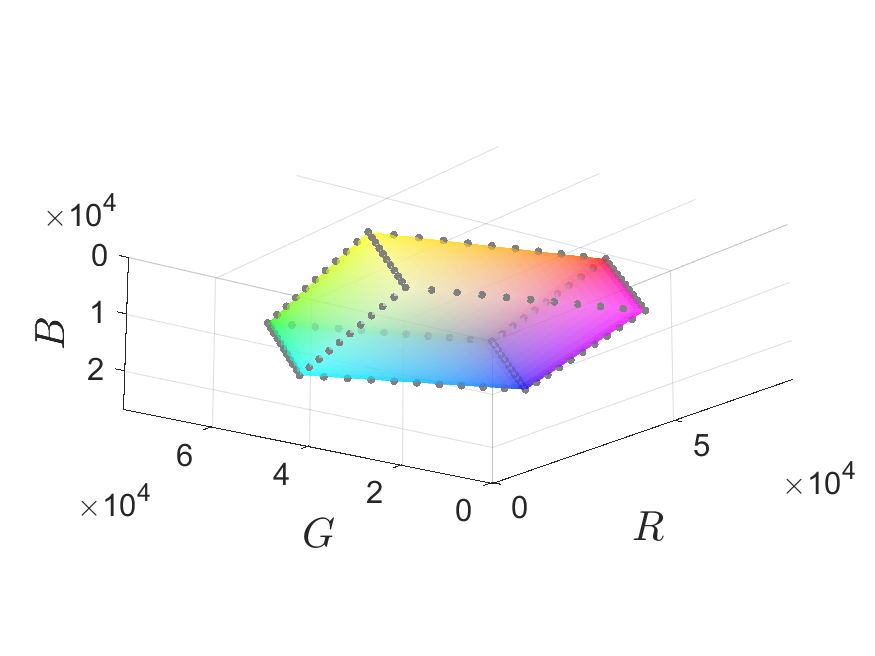}}\endminipage
  \minipage{0.33\linewidth}\center{\includegraphics[width=\linewidth]{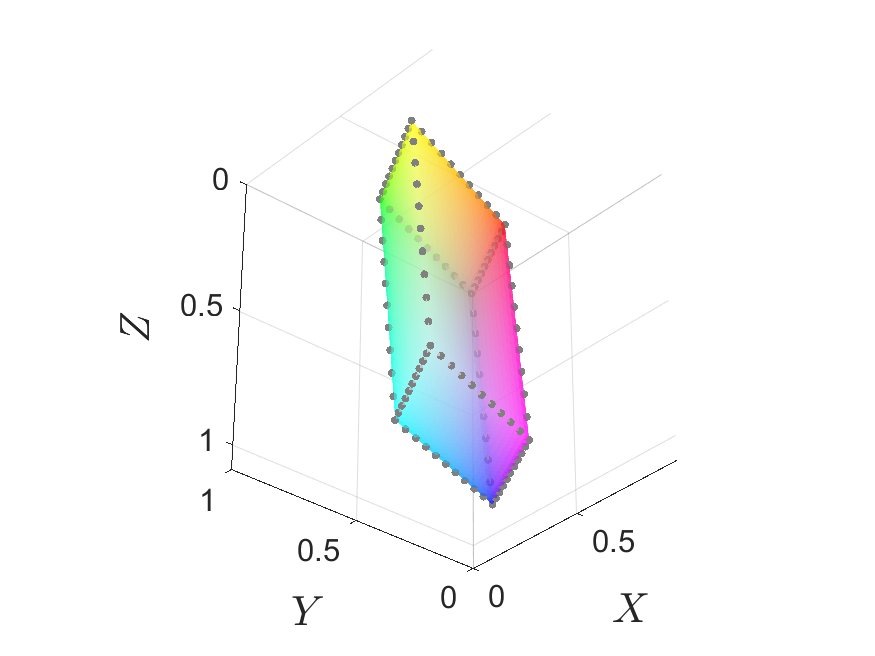}}\endminipage\vspace{1ex}
  \minipage{0.33\linewidth}\center{LMS}\endminipage
  \minipage{0.33\linewidth}\center{deviceRGB}\endminipage
  \minipage{0.33\linewidth}\center{CIE XYZ}\endminipage\vspace{2ex}

  \minipage{0.33\linewidth}\center{\includegraphics[width=\linewidth]{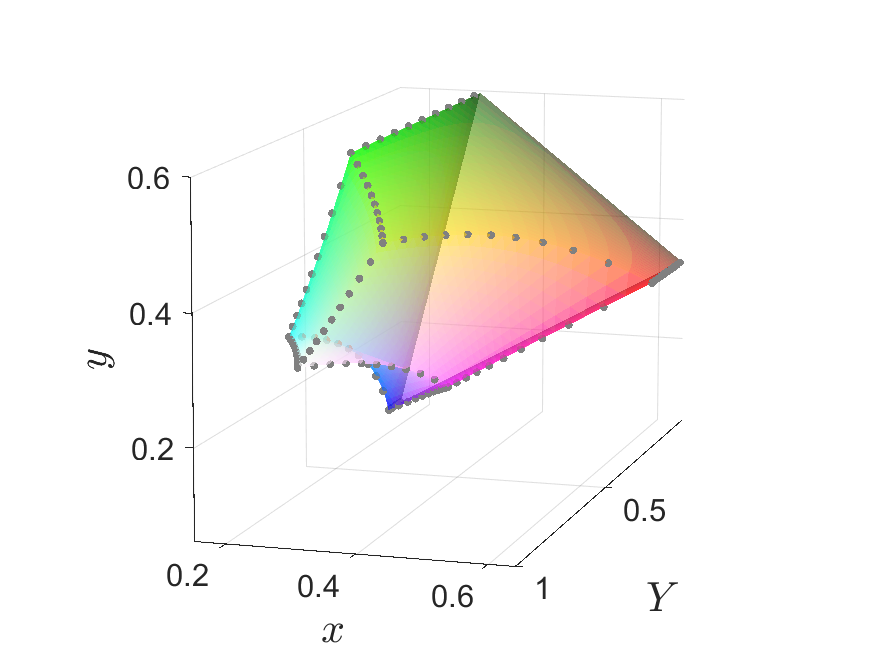}}\endminipage
  \minipage{0.33\linewidth}\center{\includegraphics[width=\linewidth]{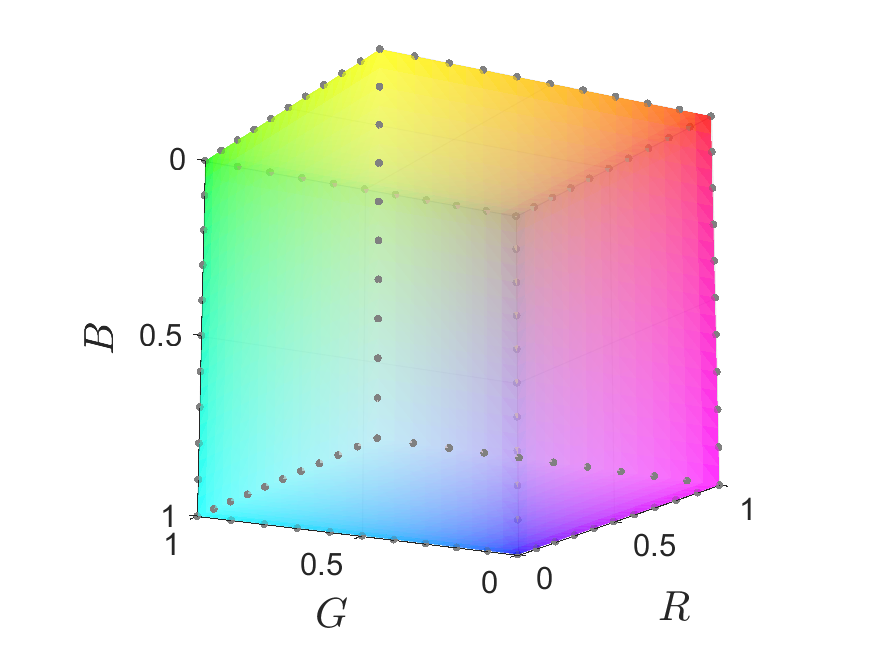}}\endminipage
  \minipage{0.33\linewidth}\center{\includegraphics[width=\linewidth]{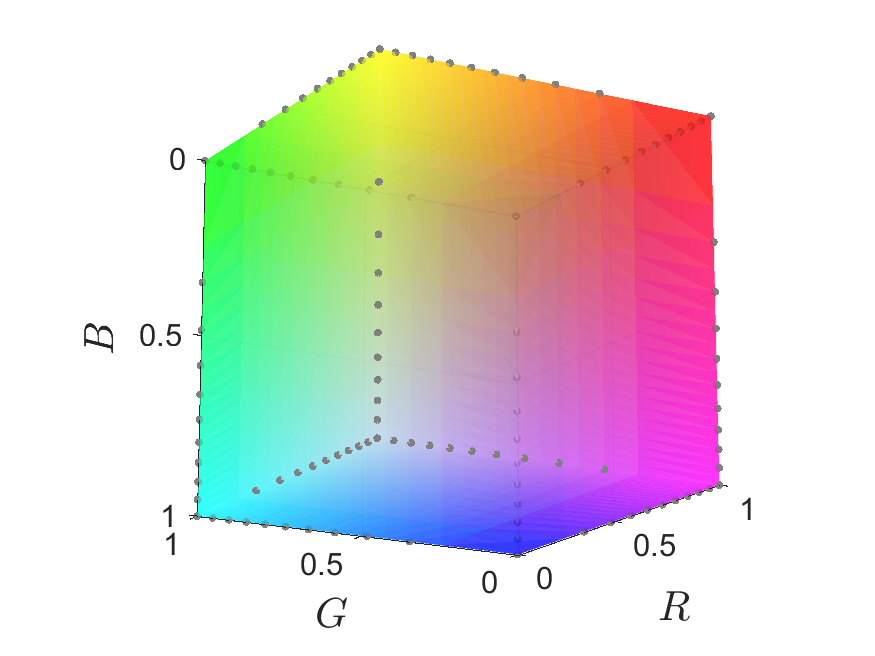}}\endminipage\vspace{1ex}
  \minipage{0.33\linewidth}\center{CIE xyY}\endminipage
  \minipage{0.33\linewidth}\center{linRGB}\endminipage
  \minipage{0.33\linewidth}\center{sRGB}\endminipage\vspace{2ex}

  \minipage{0.33\linewidth}\center{\includegraphics[width=\linewidth]{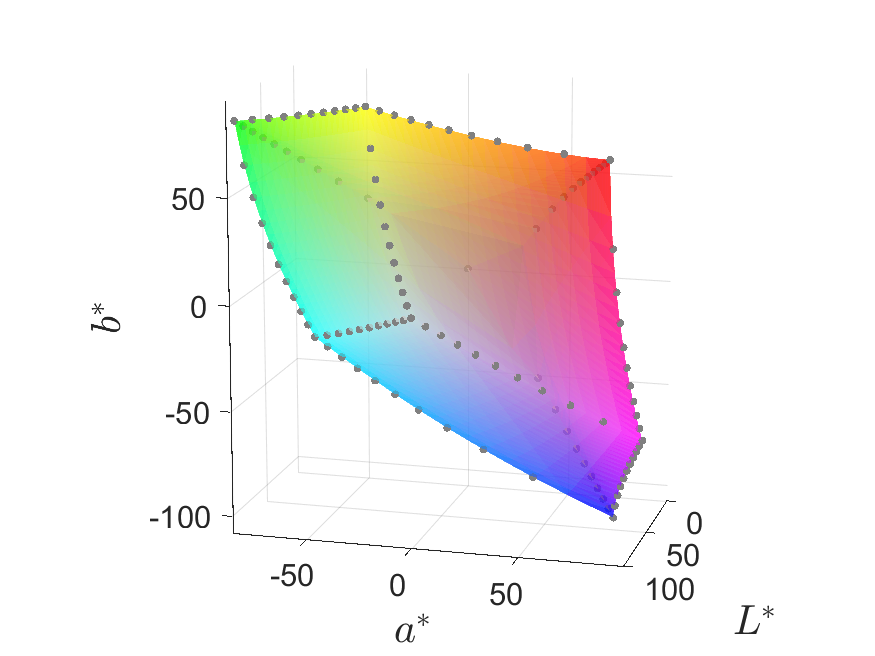}}\endminipage
  \minipage{0.33\linewidth}\center{\includegraphics[width=\linewidth]{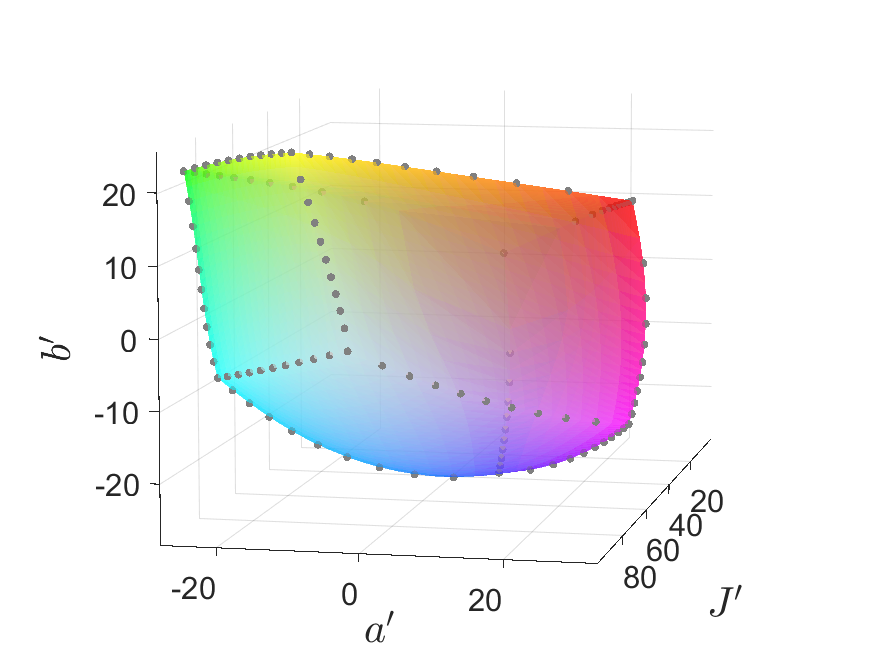}}\endminipage
  \minipage{0.33\linewidth}\center{\includegraphics[width=\linewidth]{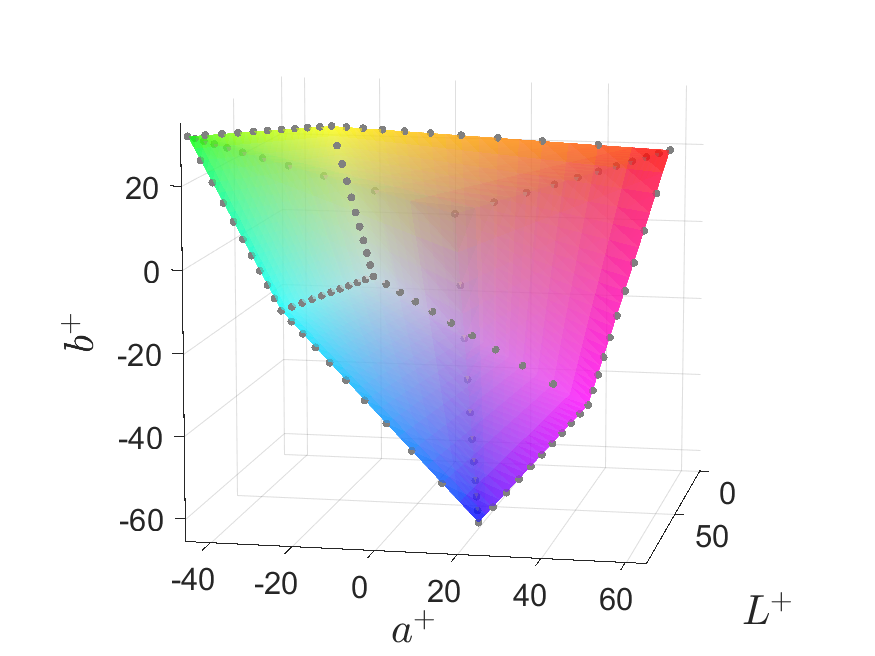}}\endminipage\vspace{1ex}
  \minipage{0.33\linewidth}\center{CIELAB}\endminipage
  \minipage{0.33\linewidth}\center{CAM16-UCS}\endminipage
  \minipage{0.33\linewidth}\center{proLab}\endminipage\vspace{2ex}

  \caption{\label{fig:sRGB_gamut}
    The sRGB gamut in various colour coordinate spaces.
  }
\end{figure*}

\begin{figure*}[t!]
  \centering
  
  \minipage{0.33\linewidth}\center{\includegraphics[width=\linewidth]{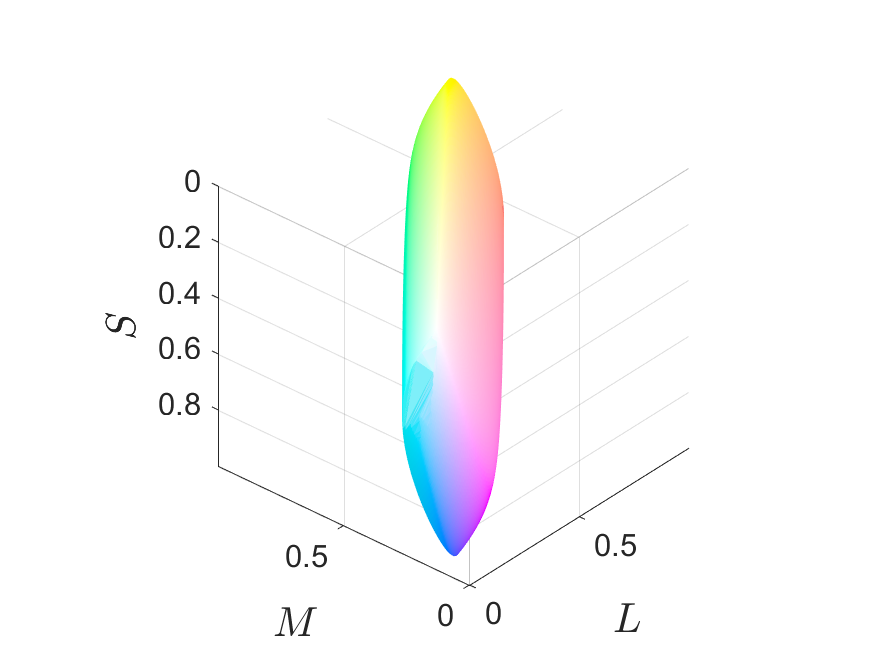}}\endminipage
  \minipage{0.33\linewidth}\center{\includegraphics[width=\linewidth]{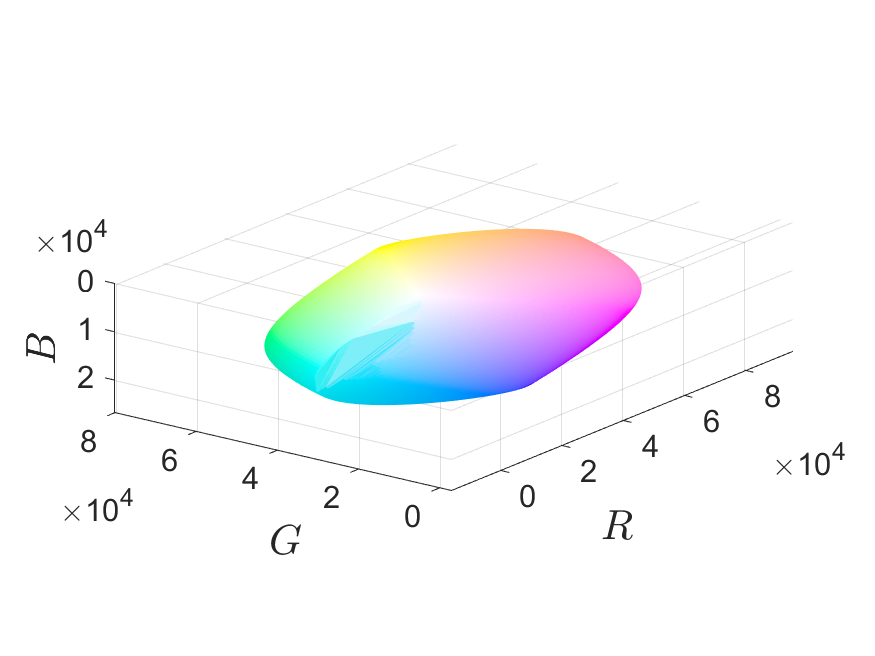}}\endminipage
  \minipage{0.33\linewidth}\center{\includegraphics[width=\linewidth]{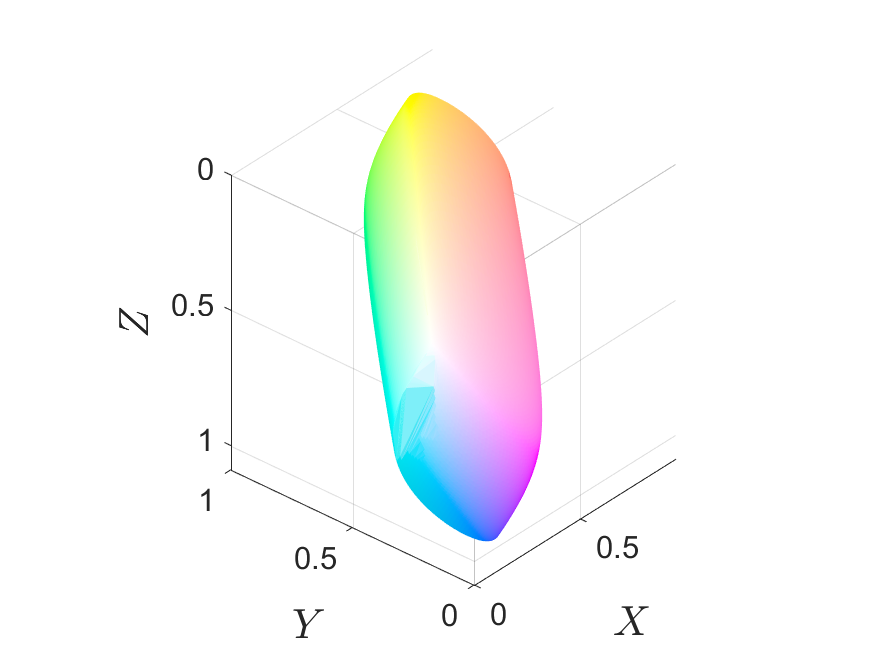}}\endminipage\vspace{1ex}
  \minipage{0.33\linewidth}\center{LMS}\endminipage
  \minipage{0.33\linewidth}\center{deviceRGB}\endminipage
  \minipage{0.33\linewidth}\center{CIE XYZ}\endminipage\vspace{2ex}

  \minipage{0.33\linewidth}\center{\includegraphics[width=\linewidth]{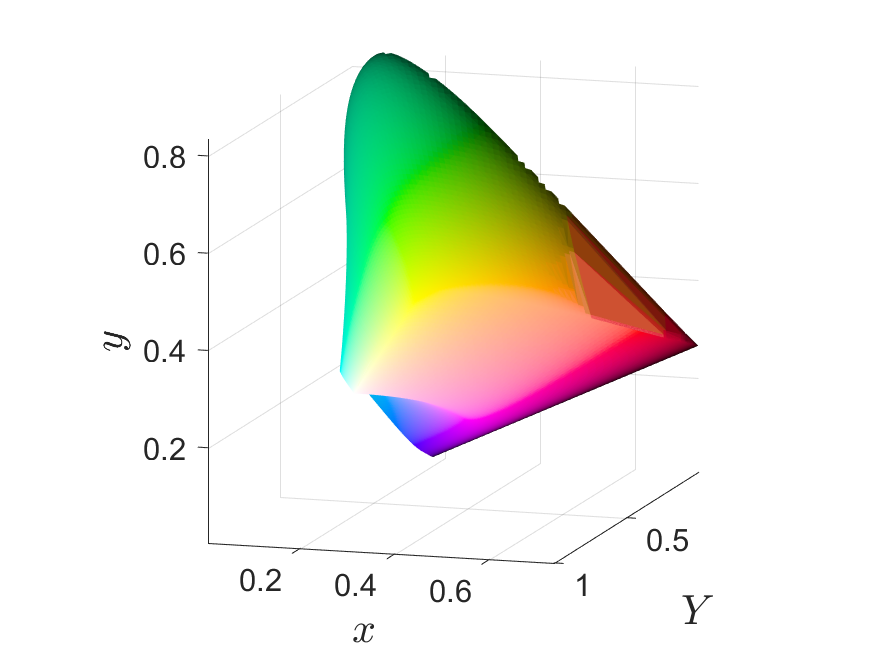}}\endminipage
  \minipage{0.33\linewidth}\center{\includegraphics[width=\linewidth]{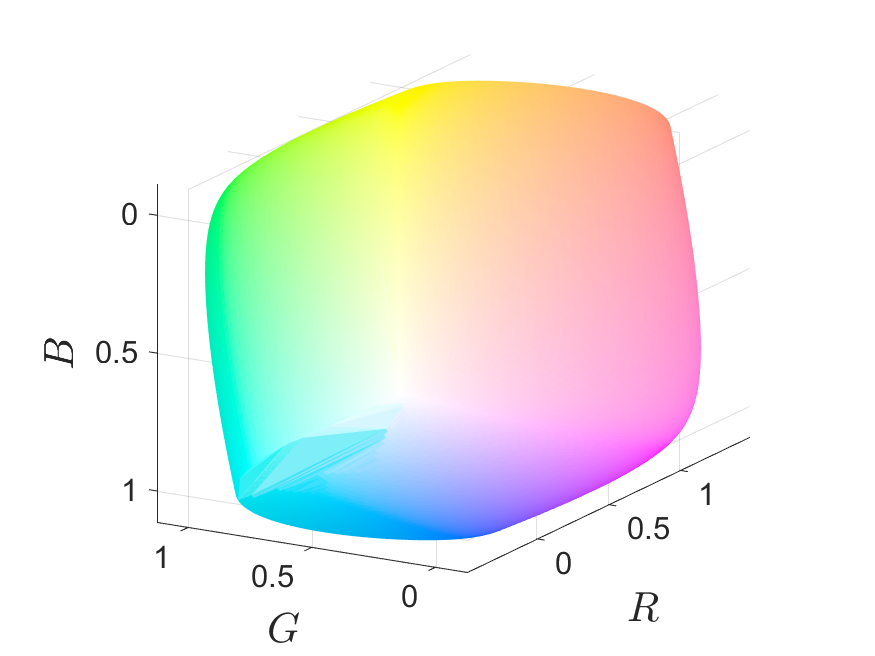}}\endminipage
  \minipage{0.33\linewidth}\center{\includegraphics[width=\linewidth]{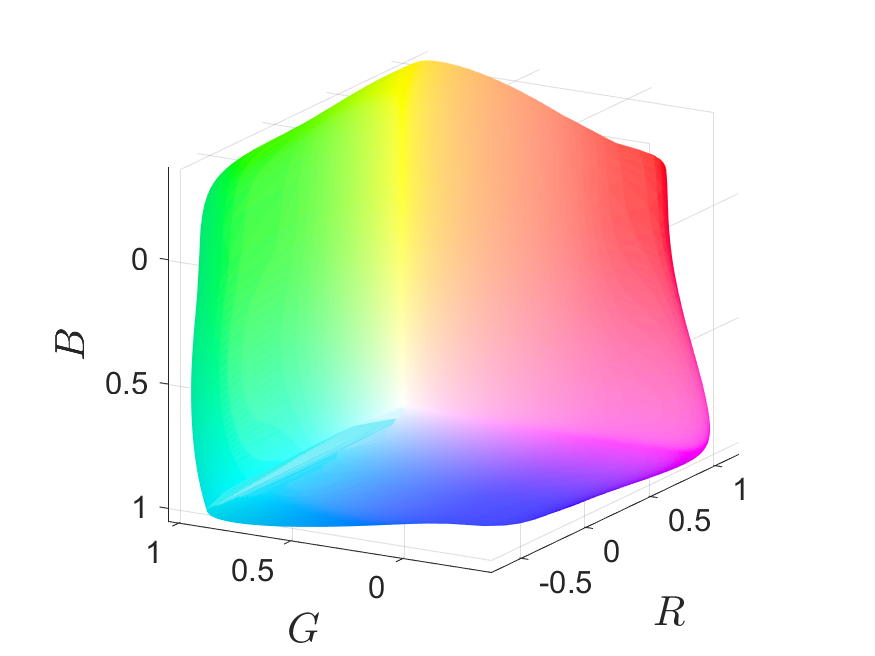}}\endminipage\vspace{1ex}
  \minipage{0.33\linewidth}\center{CIE xyY}\endminipage
  \minipage{0.33\linewidth}\center{linRGB}\endminipage
  \minipage{0.33\linewidth}\center{sRGB}\endminipage\vspace{2ex}

  \minipage{0.33\linewidth}\center{\includegraphics[width=\linewidth]{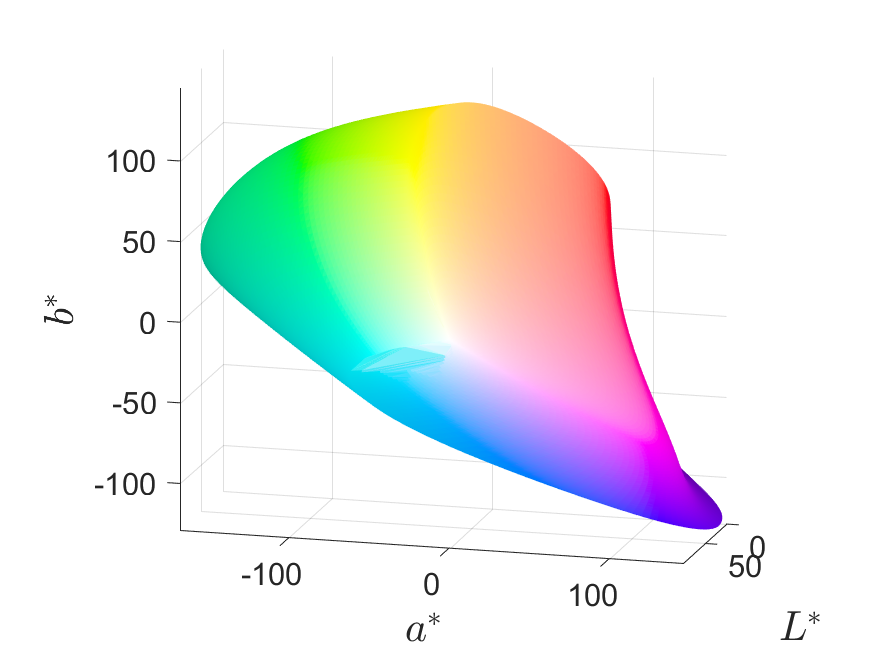}}\endminipage
  \minipage{0.33\linewidth}\center{\includegraphics[width=\linewidth]{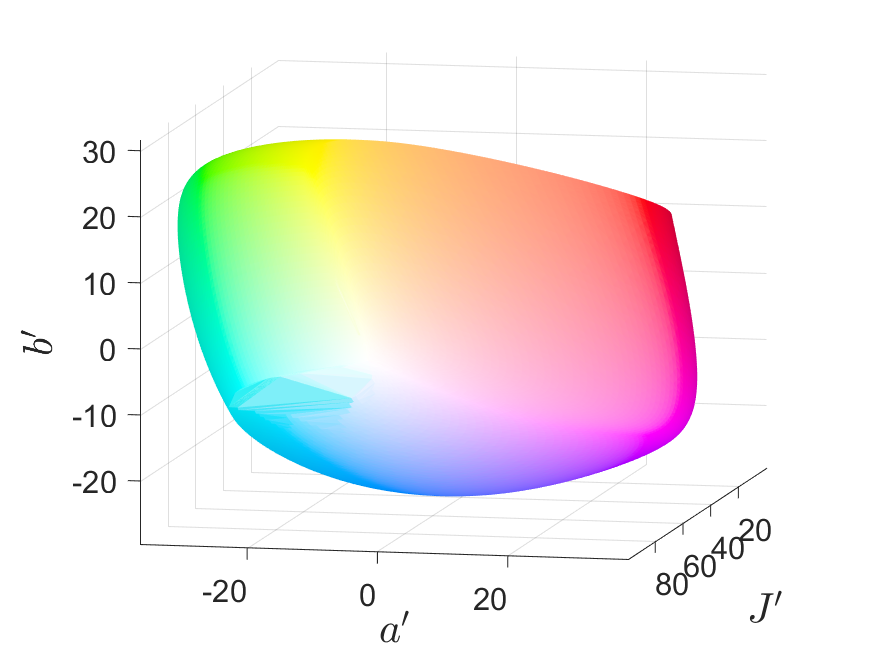}}\endminipage
  \minipage{0.33\linewidth}\center{\includegraphics[width=\linewidth]{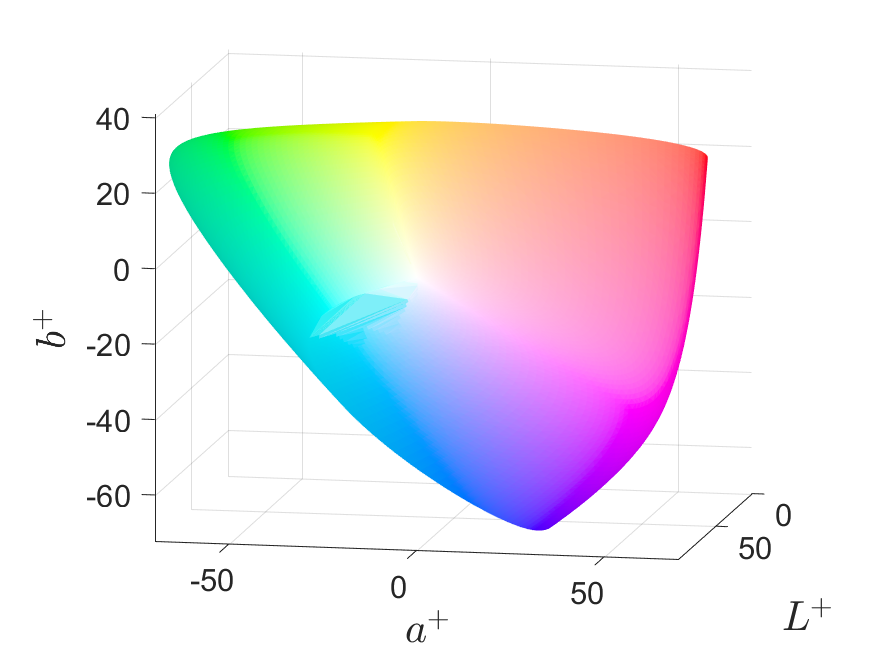}}\endminipage\vspace{1ex}
  \minipage{0.33\linewidth}\center{CIELAB}\endminipage
  \minipage{0.33\linewidth}\center{CAM16-UCS}\endminipage
  \minipage{0.33\linewidth}\center{proLab}\endminipage\vspace{2ex}

  \caption{\label{fig:color_body}
    D65 light source gamut in various colour coordinate spaces.
  }
\end{figure*}

Let us employ various visualizations to analyse the differences between colour coordinate spaces more clearly.
Figures~\ref{fig:sRGB_gamut} and~\ref{fig:color_body} show the sRGB display gamut in various colour coordinate spaces.
The saturation of colours used for the visualization was decreased significantly on this illustration due to  colour coverage restrictions.
ProLab preserves the shape of the sRGB gamut as a hexahedron.
Another advantage is that proLab, unlike CIELAB, keeps the convexity of the gamut.

\begin{figure*}[t!]
  \textbf{\centering}
  
  \minipage{0.33\linewidth}\center{\includegraphics[width=\linewidth]{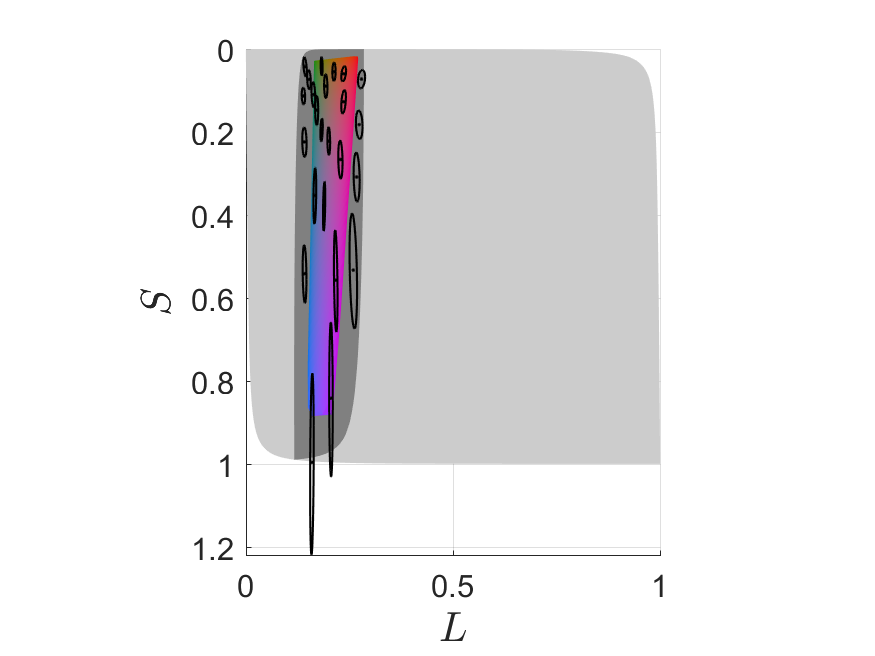}}\endminipage
  \minipage{0.33\linewidth}\center{\includegraphics[width=\linewidth]{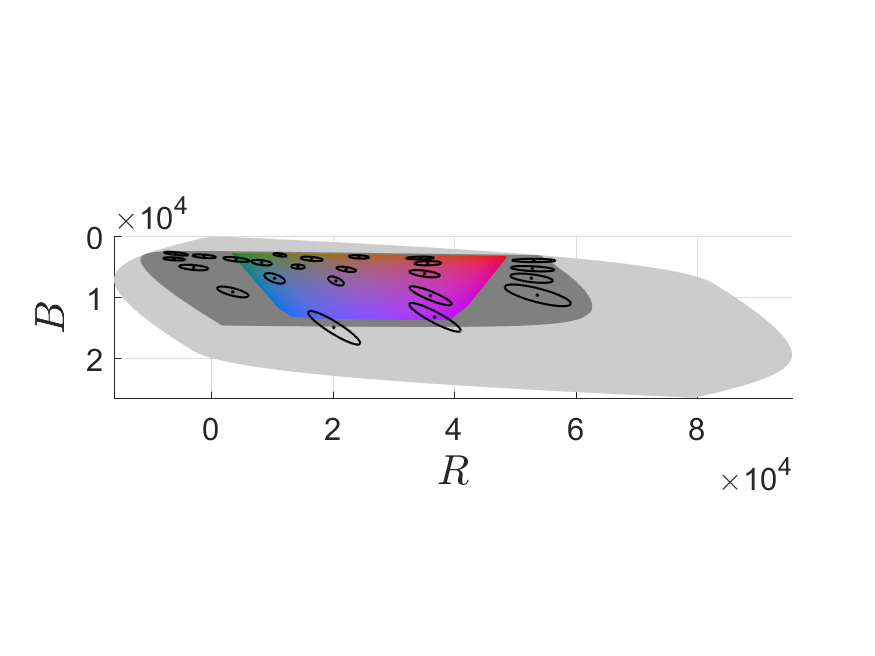}}\endminipage
  \minipage{0.33\linewidth}\center{\includegraphics[width=\linewidth]{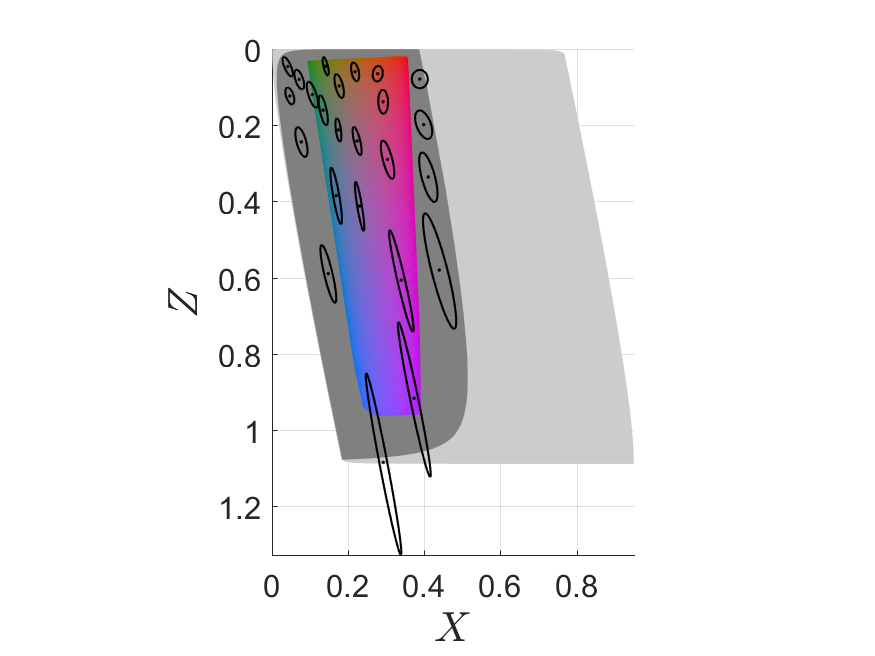}}\endminipage\vspace{1ex}
  \minipage{0.33\linewidth}\center{LMS}\endminipage
  \minipage{0.33\linewidth}\center{deviceRGB}\endminipage
  \minipage{0.33\linewidth}\center{CIE XYZ}\endminipage\vspace{2ex}

  \minipage{0.33\linewidth}\center{\includegraphics[width=\linewidth]{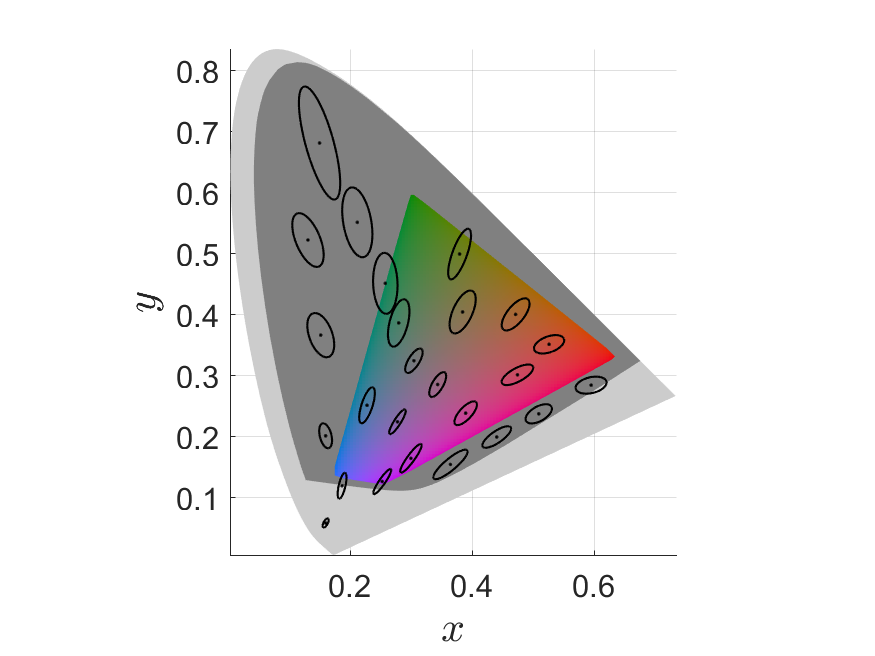}}\endminipage
  \minipage{0.33\linewidth}\center{\includegraphics[width=\linewidth]{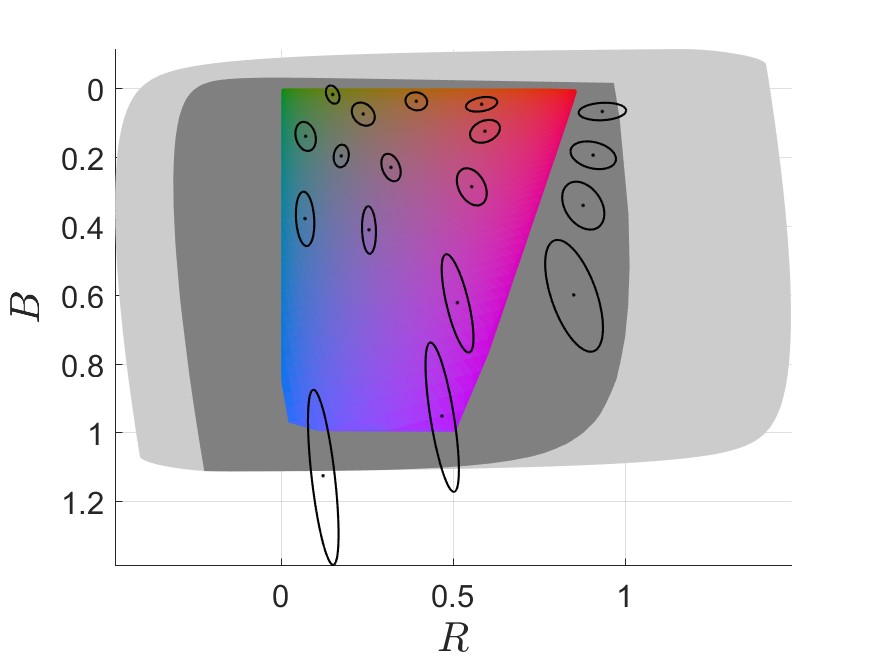}}\endminipage
  \minipage{0.33\linewidth}\center{\includegraphics[width=\linewidth]{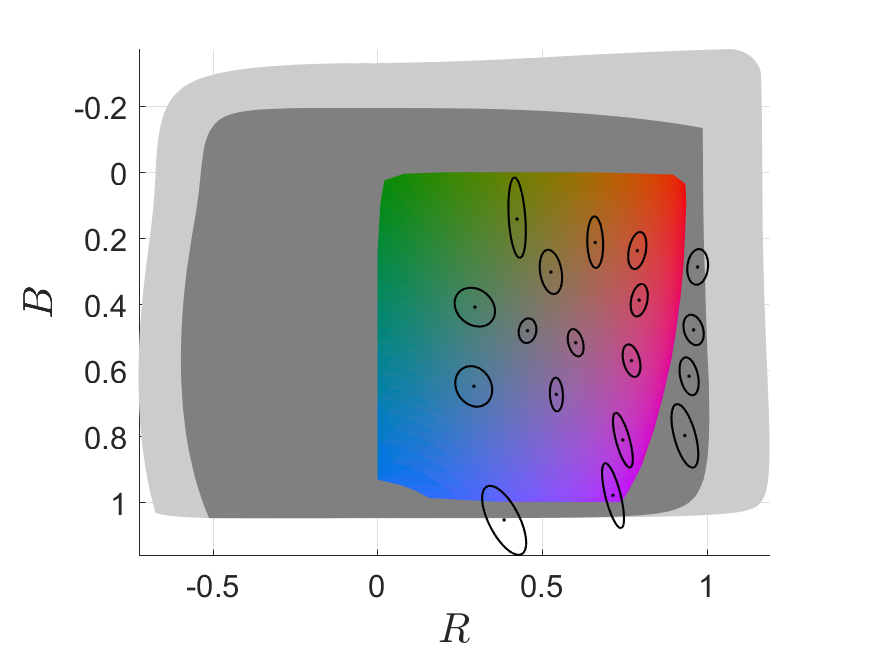}}\endminipage\vspace{1ex}
  \minipage{0.33\linewidth}\center{CIE xyY}\endminipage
  \minipage{0.33\linewidth}\center{linRGB}\endminipage
  \minipage{0.33\linewidth}\center{sRGB}\endminipage\vspace{2ex}

  \minipage{0.33\linewidth}\center{\includegraphics[width=\linewidth]{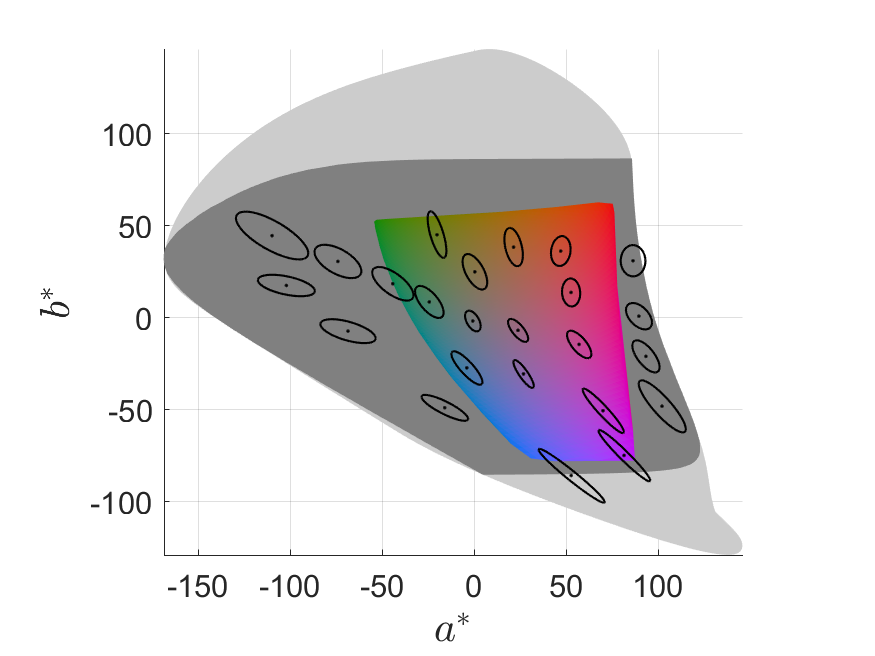}}\endminipage
  \minipage{0.33\linewidth}\center{\includegraphics[width=\linewidth]{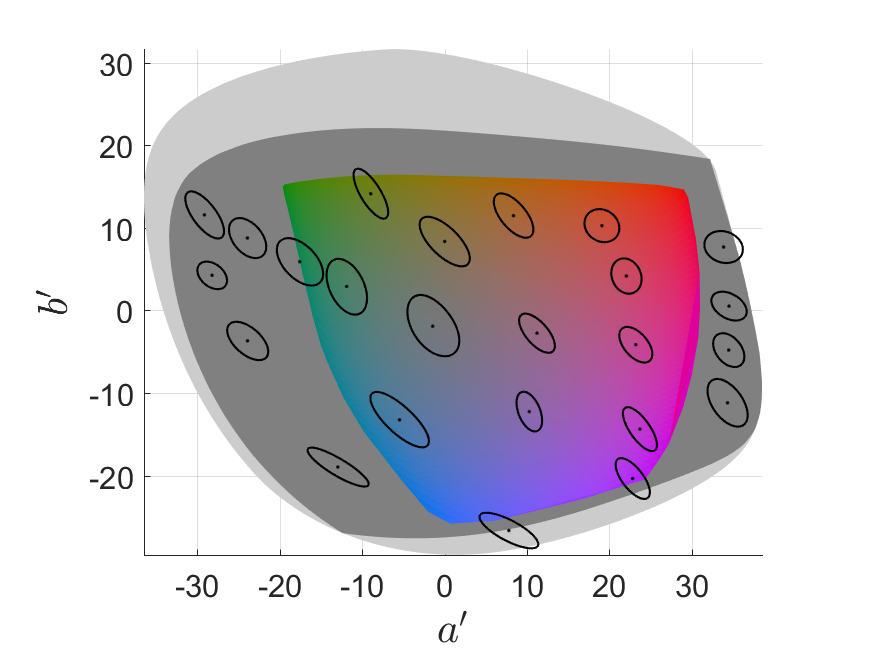}}\endminipage
  \minipage{0.33\linewidth}\center{\includegraphics[width=\linewidth]{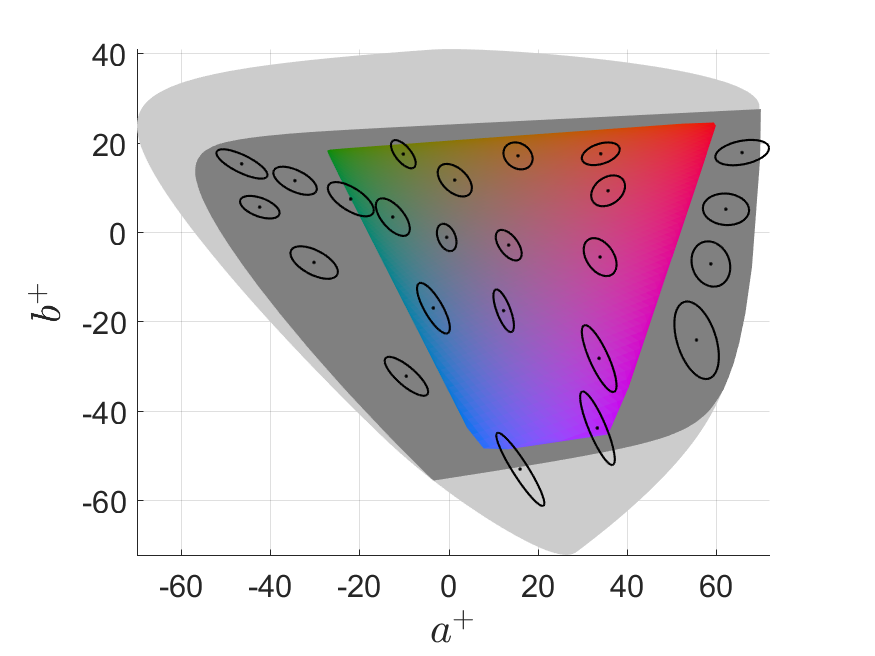}}\endminipage\vspace{1ex}
  \minipage{0.33\linewidth}\center{CIELAB}\endminipage
  \minipage{0.33\linewidth}\center{CAM16-UCS}\endminipage
  \minipage{0.33\linewidth}\center{proLab}\endminipage\vspace{2ex}

  \caption{\label{fig:macadam}
    MacAdam ellipses for $L^*=50$. The sRGB display gamut cross-section is shown in rainbow colours; D65 light source gamut cross-section and projection are shown with dark grey and light grey, respectively.
  }
\end{figure*}

In Figure~\ref{fig:macadam}, we show the colour non-uniformity over  chromaticity diagram with MacAdam ellipses~\cite{macadam1942visual}, which are just noticeable colour differences (JND) scaled up 10 times; they were originally defined in CIE xyY.
To plot them for each given colour space $C_\Phi$, we use a linear approximation of the transformation $\Phi$ around the ellipse centres.

\begin{figure*}[t!]
  \centering
  
  \minipage{0.33\linewidth}\center{\includegraphics[width=\linewidth]{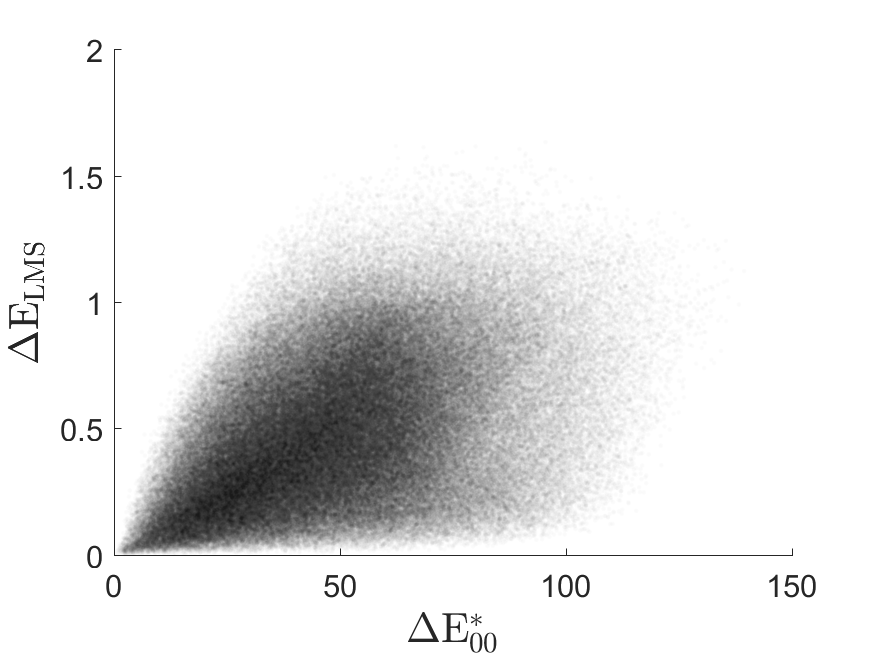}}\endminipage
  \minipage{0.33\linewidth}\center{\includegraphics[width=\linewidth]{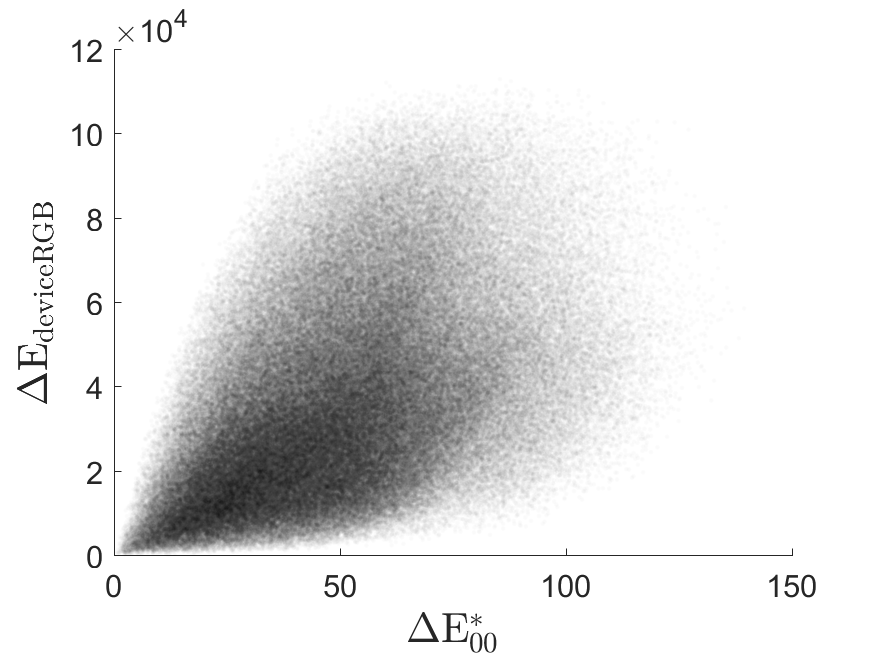}}\endminipage
  \minipage{0.33\linewidth}\center{\includegraphics[width=\linewidth]{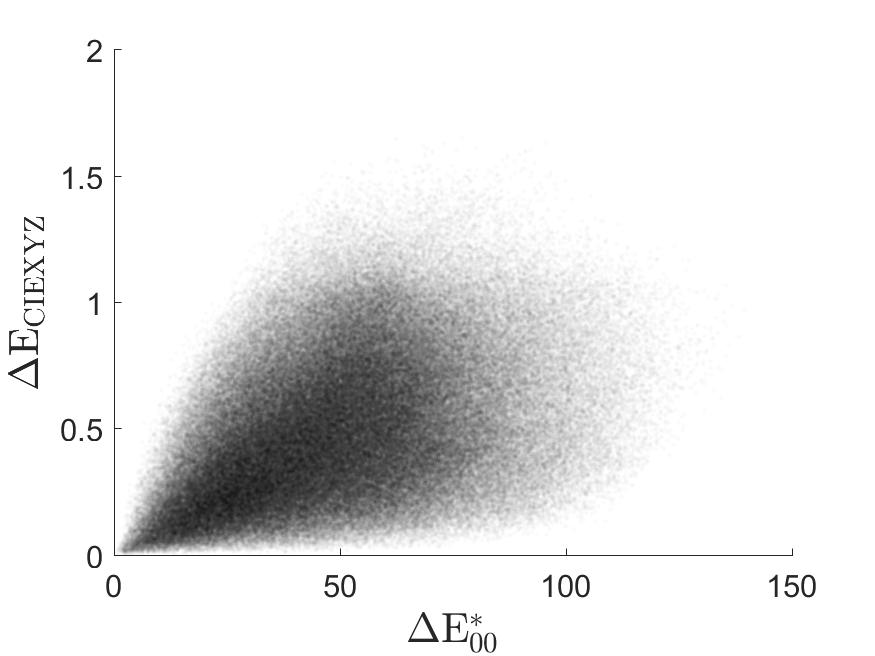}}\endminipage\vspace{1ex}
  \minipage{0.33\linewidth}\center{LMS}\endminipage
  \minipage{0.33\linewidth}\center{deviceRGB}\endminipage
  \minipage{0.33\linewidth}\center{CIE XYZ}\endminipage\vspace{2ex}

  \minipage{0.33\linewidth}\center{\includegraphics[width=\linewidth]{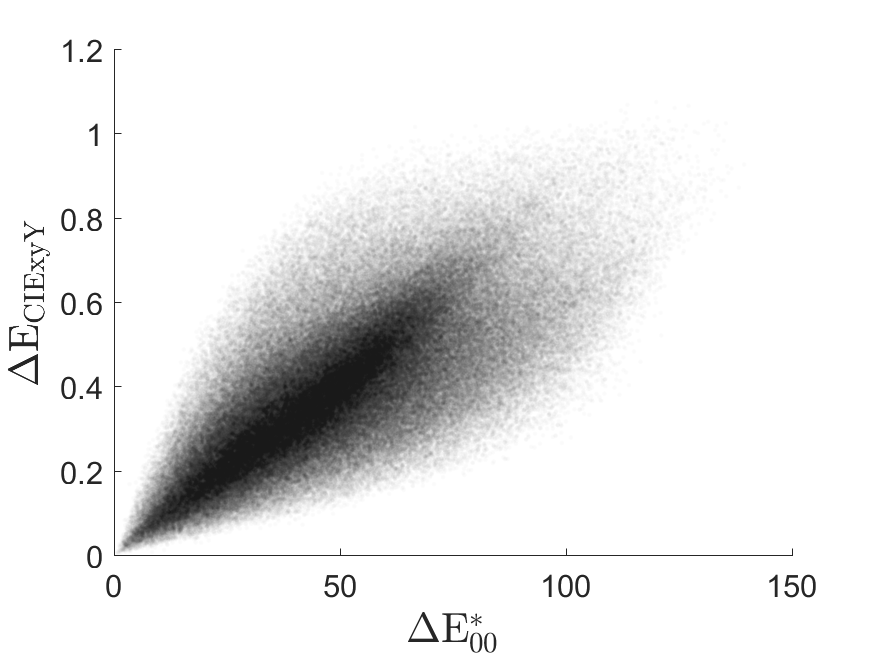}}\endminipage
  \minipage{0.33\linewidth}\center{\includegraphics[width=\linewidth]{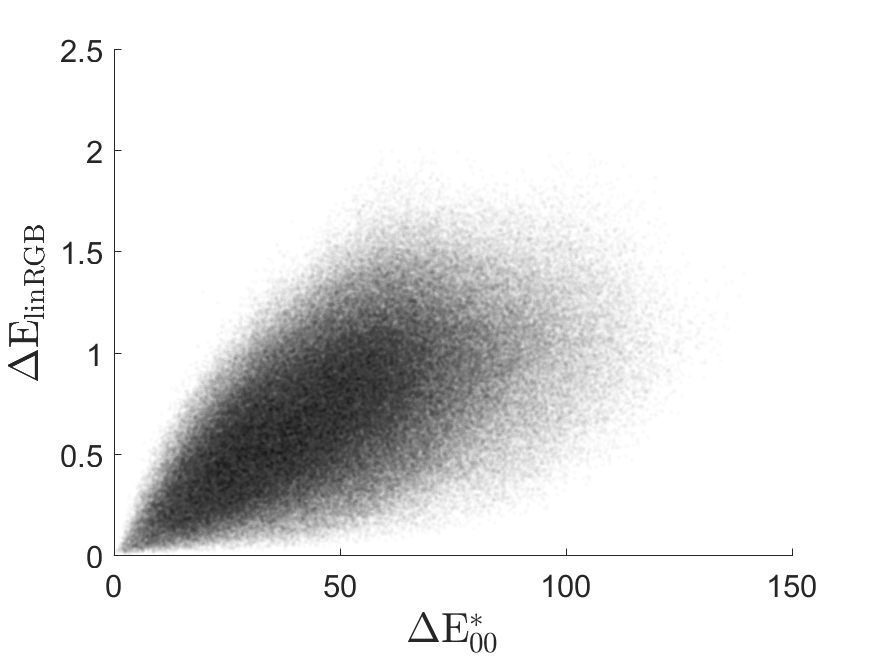}}\endminipage
  \minipage{0.33\linewidth}\center{\includegraphics[width=\linewidth]{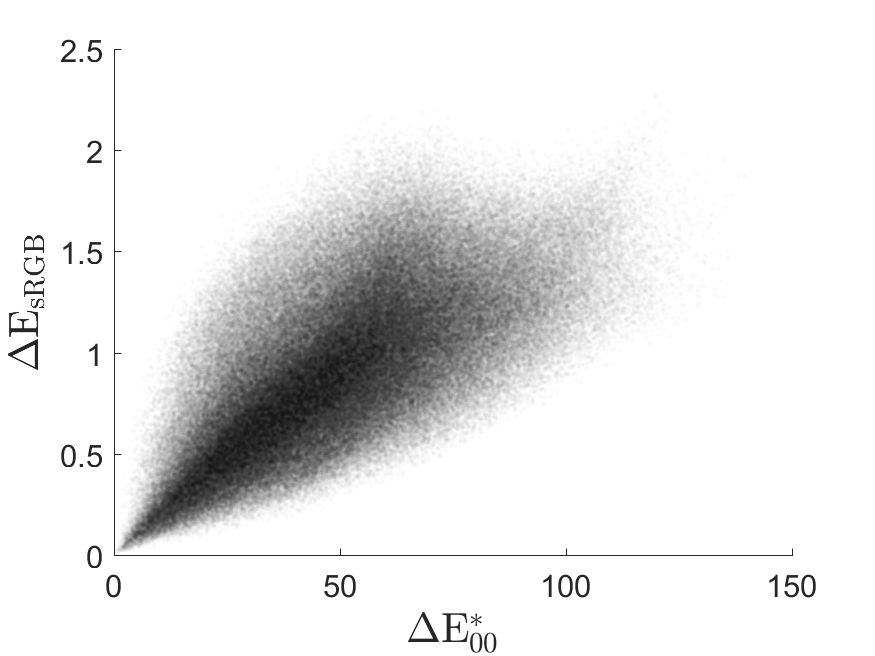}}\endminipage\vspace{1ex}
  \minipage{0.33\linewidth}\center{CIE xyY}\endminipage
  \minipage{0.33\linewidth}\center{linRGB}\endminipage
  \minipage{0.33\linewidth}\center{sRGB}\endminipage\vspace{2ex}

  \minipage{0.33\linewidth}\center{\includegraphics[width=\linewidth]{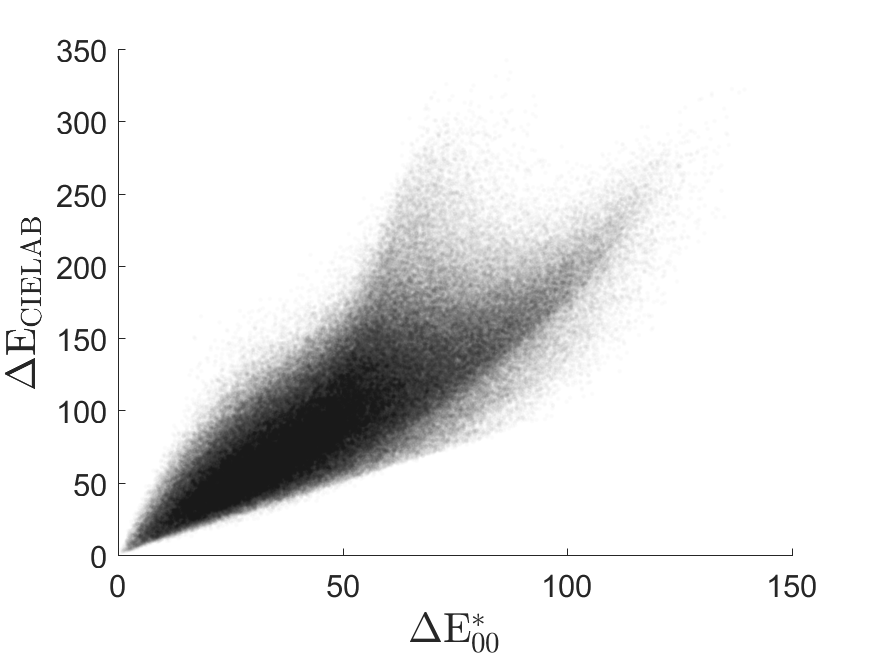}}\endminipage
  \minipage{0.33\linewidth}\center{\includegraphics[width=\linewidth]{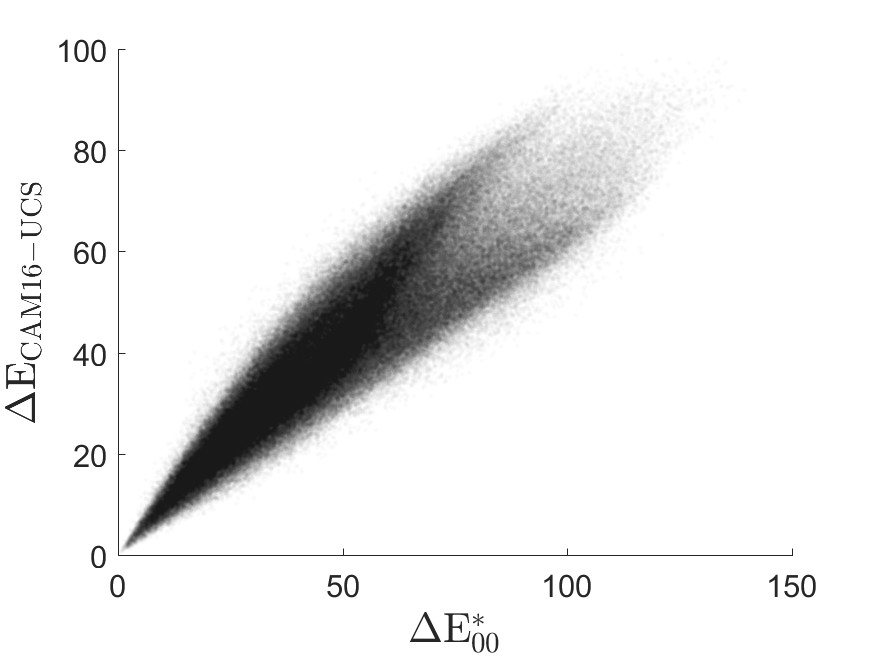}}\endminipage
  \minipage{0.33\linewidth}\center{\includegraphics[width=\linewidth]{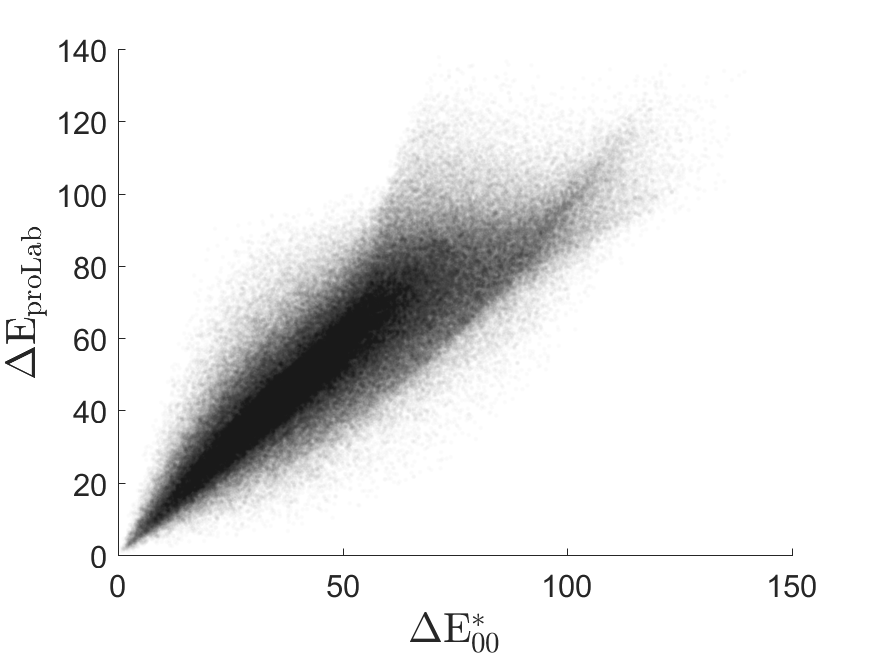}}\endminipage\vspace{1ex}
  \minipage{0.33\linewidth}\center{CIELAB}\endminipage
  \minipage{0.33\linewidth}\center{CAM16-UCS}\endminipage
  \minipage{0.33\linewidth}\center{proLab}\endminipage\vspace{2ex}

  \caption{\label{fig:scatter}
    Joint distributions of  CIEDE2000 colour differences $\func{{\Delta E}^*_{00}}$ and Euclidean distances $\func{\Delta E_\Phi}$ for various colour coordinate systems.
  }
\end{figure*}

In Fig.~\ref{fig:scatter}, we visualize colour non-uniformities with joint distributions of Euclidian distances $\func{\Delta E_\Phi}$ and CIEDE2000 colour differences $\func{{\Delta E}^*_{00}}$ over test sample $G^2_{n_2}$ for each colour coordinate space.
On such scatter plots, the thinner the cluster along the line passing through $\vect{0}$, the more perceptually uniform the colour coordinate space is.
CAM16-UCS and proLab joint distributions have significantly better shapes than those of the other colour coordinate spaces,
but proLab is inferior to CAM16-UCS in the middle range distances.
Note that in a region of large Euclidean distances, CAM16-UCS has two distinctly different loci in such region, which means there is  significant non-uniformity in this range.
Strongly decorrelated regions similar in location and shape are observed in proLab and CIELAB; however, for CIELAB the region is larger and deviates more and farther from the main locus.

\begin{figure*}[t!]
  \centering
  
  \minipage{0.33\linewidth}\center{\includegraphics[width=\linewidth]{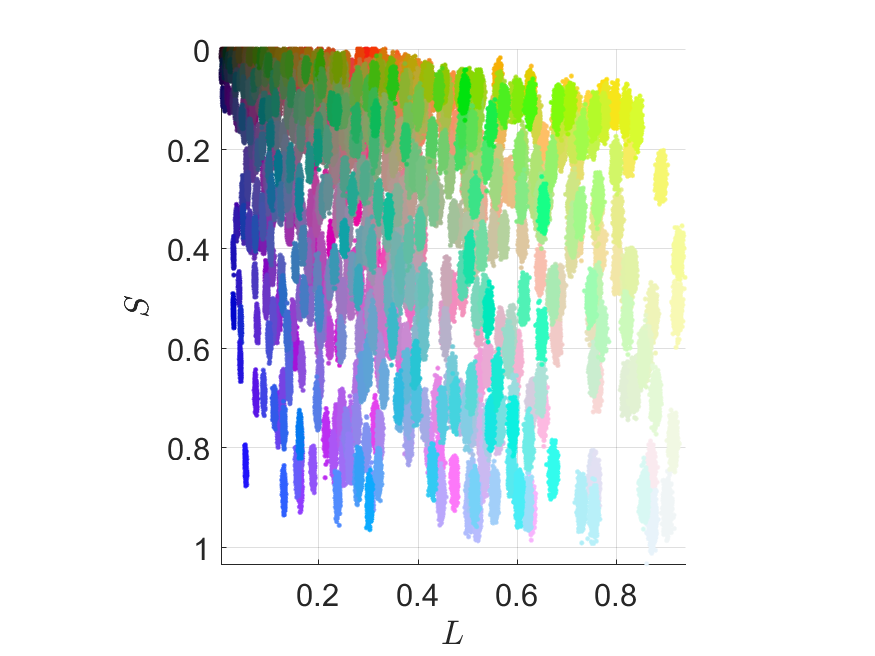}}\endminipage
  \minipage{0.33\linewidth}\center{\includegraphics[width=\linewidth]{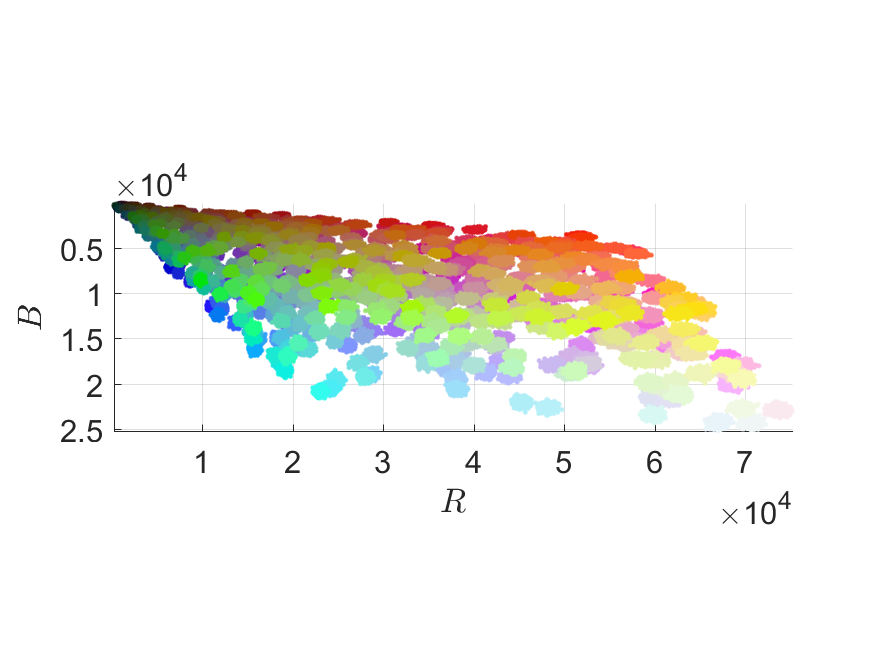}}\endminipage
  \minipage{0.33\linewidth}\center{\includegraphics[width=\linewidth]{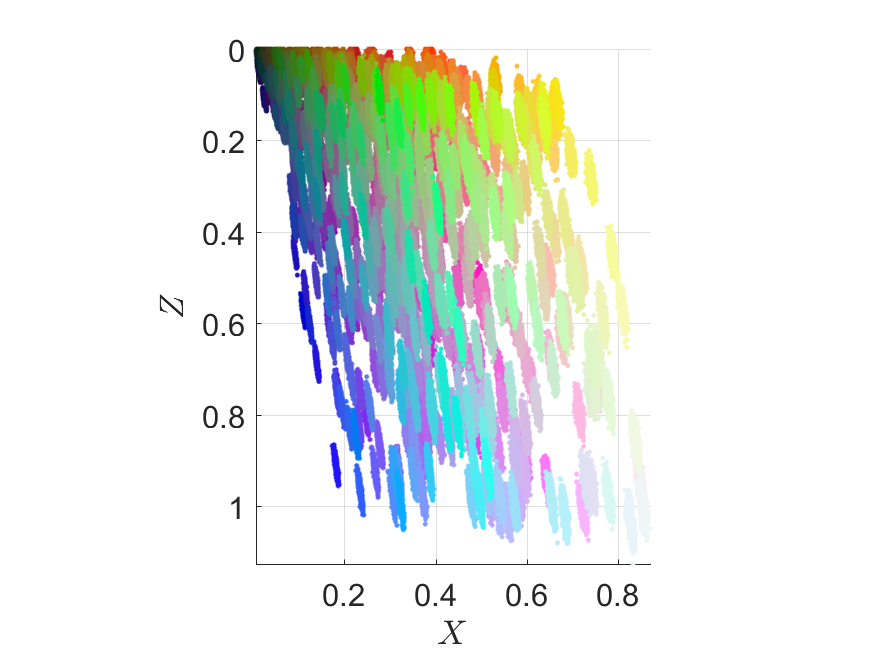}}\endminipage\vspace{1ex}
  \minipage{0.33\linewidth}\center{LMS}\endminipage
  \minipage{0.33\linewidth}\center{deviceRGB}\endminipage
  \minipage{0.33\linewidth}\center{CIE XYZ}\endminipage\vspace{2ex}

  \minipage{0.33\linewidth}\center{\includegraphics[width=\linewidth]{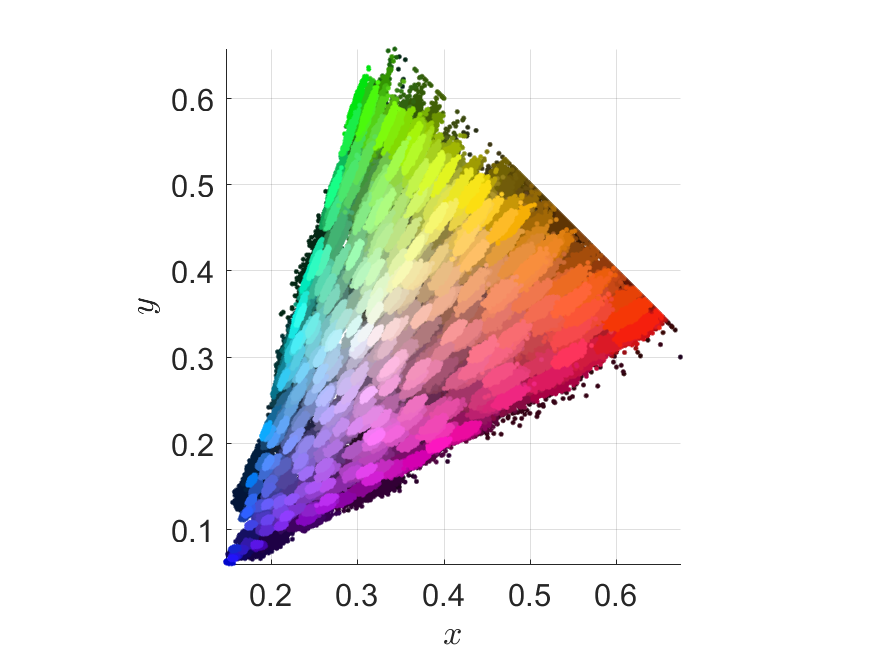}}\endminipage
  \minipage{0.33\linewidth}\center{\includegraphics[width=\linewidth]{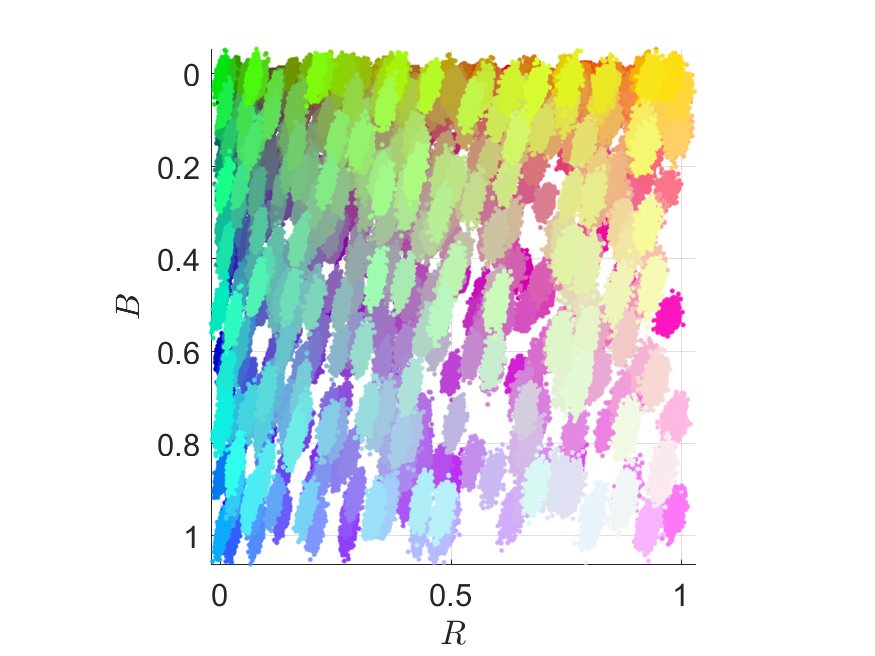}}\endminipage
  \minipage{0.33\linewidth}\center{\includegraphics[width=\linewidth]{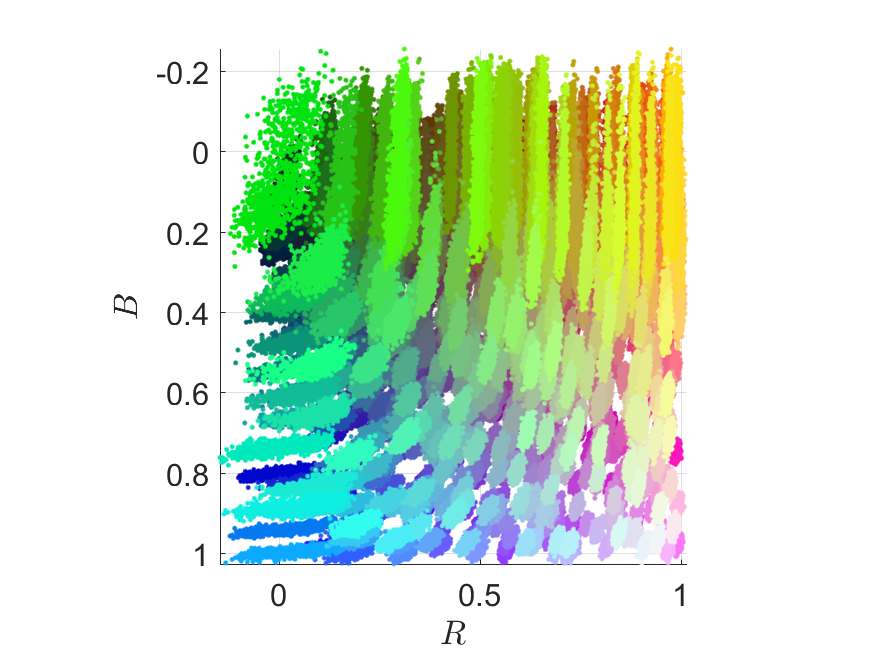}}\endminipage\vspace{1ex}
  \minipage{0.33\linewidth}\center{CIE xyY}\endminipage
  \minipage{0.33\linewidth}\center{linRGB}\endminipage
  \minipage{0.33\linewidth}\center{sRGB}\endminipage\vspace{2ex}

  \minipage{0.33\linewidth}\center{\includegraphics[width=\linewidth]{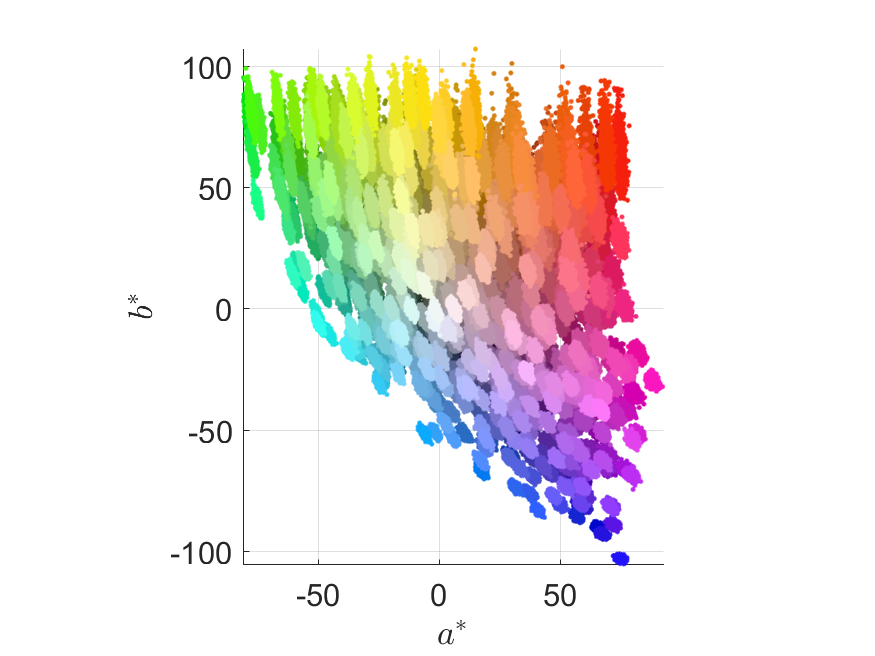}}\endminipage
  \minipage{0.33\linewidth}\center{\includegraphics[width=\linewidth]{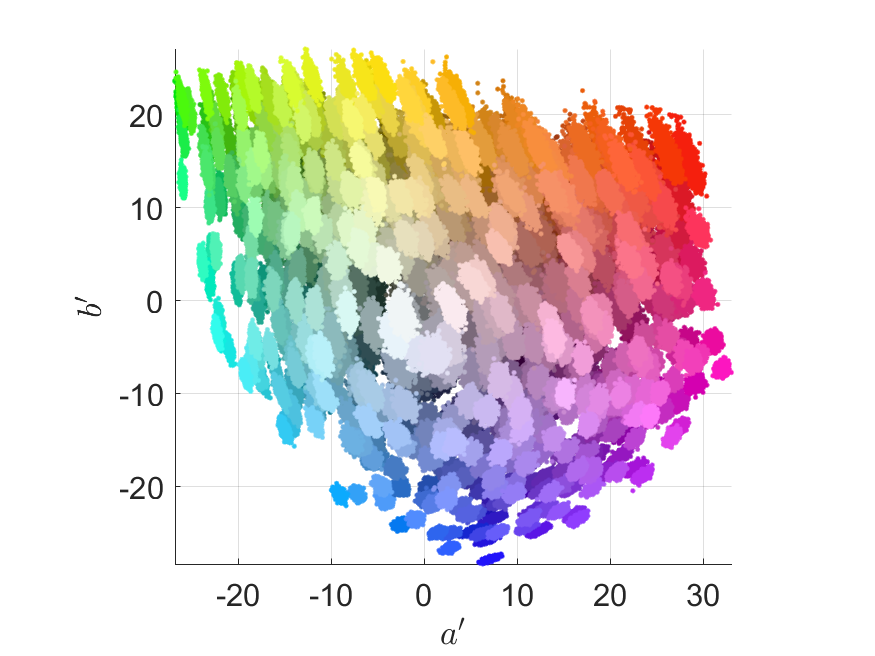}}\endminipage
  \minipage{0.33\linewidth}\center{\includegraphics[width=\linewidth]{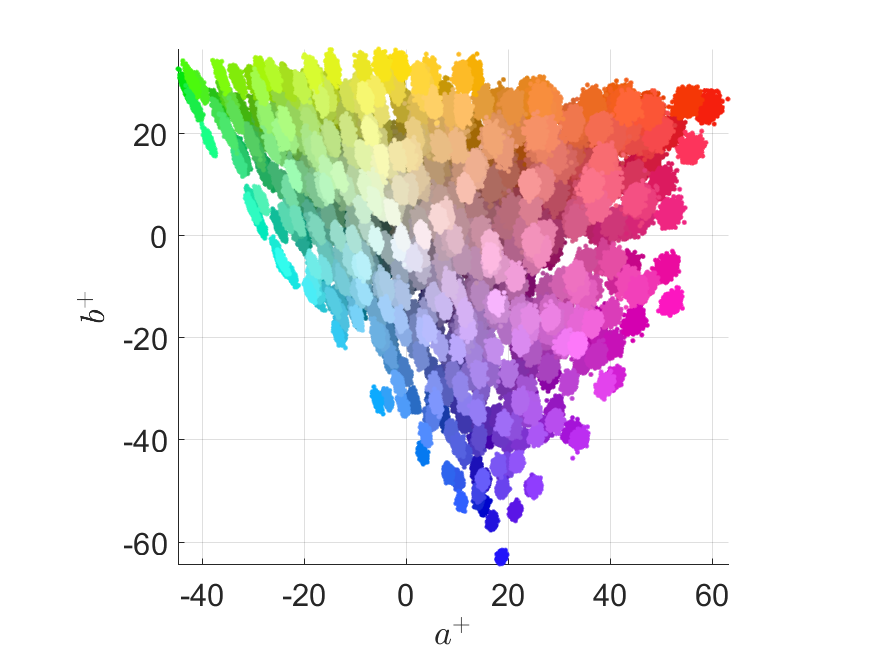}}\endminipage\vspace{1ex}
  \minipage{0.33\linewidth}\center{CIELAB}\endminipage
  \minipage{0.33\linewidth}\center{CAM16-UCS}\endminipage
  \minipage{0.33\linewidth}\center{proLab}\endminipage\vspace{2ex}

  \caption{\label{fig:noise_clouds}
    Visualization of sensor noise in different colour coordinate spaces within the sRGB display gamut.
    Each individual cloud is a projection of a certain  measured colour sample for which the noise is described by the J\"{a}hne model.
  }
\end{figure*}

To provide detailed visualization of noise heteroscedasticity, we plot charts similar to MacAdam ellipses (Fig.~\ref{fig:noise_clouds}).
For a set of colours from the sRGB gamut, we model the noise distributions of measurements according to the parameters of the model~\eqref{eq:Yane_noise_model_special}.
Each distribution is projected onto the given colour coordinate space and visualized with its averaged colour.
The three-dimensional structure of noise parameters is visualized by plotting the lighter colours over the dark ones.

\section{Discussion}
A key feature of the proLab design is its projectivity.
Within this colour coordinate space, the central projection on any plane bypassing $\vect{0}$ is a chromaticity diagram, since proLab does not shift the coordinate origin.
That is,  colours that differ only in brightness in the original space of spectral irradiance are mapped into a single point.

Another interesting property of proLab concerns  image shot noise.
In a noisy image, colour estimation by arithmetic averaging is valid only in linear colour spaces.
However, chromaticity estimation by linear regression is invalid even in a linear space, since the amplitude of shot noise depends on the value of the colour coordinates.
But in proLab, the linear regression procedure appears to be more correct, due to the noise being more homoscedastic compared to  standard linear spaces.

In this work, proLab was constructed for the D65 light source, so   the question arises of how to adapt these colour coordinates to a different kind of illumination.
In order to achieve the best possible accuracy, we should form sample  $G^2_{n_1}$ of colour pairs over the given light source gamut, and then optimize matrix $Q$ on this sample.
But such a procedure is inconvenient and time-consuming, so we suggest using an approach similar to CIELAB: parametrization of the transformation with the light source coordinates, while `the kernel' of the transformation $Q$ is kept the same.
We find such an approach to be optimal for proLab as well as for CIELAB, since for both systems the inaccuracy in uniformity is too significant to be fixed by a separate optimization.
That is why for  light sources other than D65, we suggest  using the same elements of the matrix $Q$ as in~\eqref{eq:Q_final}.

For both techniques, the matrix $P$ is finally defined in accordance with the von Kries adaptation model~\eqref{eq:PQ_adaptation}.
Here we use this simplification just for compatibility with  existing solutions, taking into account that the given model is criticized for poor accuracy.
Currently, several more accurate adaptation models also named after von Kries are known, which are expressed via linear transformation of colour coordinates~\cite{bianco2010two}.
Such models could also be used in proLab, since the replacement of the adaptation model by another linear (and even projective) one does not affect  matrix $Q$, and requires only redefinition of matrix $N$.

Depending on the specific task, the metric parameters of proLab could also be modified.
Particularly, it is not obvious that pairs with different colour differences ${\Delta E}^*_{00}$ should be equally weighted while solving the optimization problem.
In some possible applications large (or small) colour differences could be negligible.
In such cases, proLab parameters must be optimized with the same method but on a different $G^2_{n_1}$ sample.

Besides this, the relationship of the $L^+$ axis with brightness can be weakened in order to increase or toughen the perceptual uniformity of the result.
Alternatively, the condition~\eqref{eq:brightness_sanity} could be more strict for LMS colour coordinates.
Note that the fulfilment of requirement~\eqref{eq:brightness_sanity} directly implies the fulfilment of similar requirements on the linRGB coordinates.
All elements of transition matrix from linRGB to CIE XYZ are non-negative, and
\begin{equation}
  \vect{x} \in \R_{\ge 0}^n, \ A \in \R_{\ge 0}^{n \times n} \Longrightarrow A\vect{x} \in \R_{\ge 0}^n,
\end{equation}
i.e., a non-negative increment in linRGB implies a non-negative increment in CIE XYZ.
Since the transformation from sRGB to linRGB is component-wise monotonic, the aforesaid also implies that the lightness component is non-decreasing when increasing the sRGB coordinates.
However, in LMS space the same behaviour is not guaranteed as the transformation matrix from LMS to CIE XYZ contains negative elements.

It is important to further study the parameters of noise in various spaces, including proLab.
Our interest is focused on the experimental data for various cameras as well as on analytical models for estimation of heteroscedasticity under various conditions.
Along with  shot noise, these models could also consider sensor signal discreteness.
Outside the context of uniform colour spaces, the effects of colour digitization in technical systems have already been studied in~\cite{palchikova2018quantization}.

We also note that the secondary locus of the joint distance chart shown in Fig.~\Ref{fig:scatter} raises the question about the localization of the significantly non-projective parts of the gamut.
On the other hand, the strict requirement of projectivity was introduced formally.
In practice, the regression errors caused by the non-projectivity of the model may turn out to be insignificant compared to the noise.
Therefore, further study could also involve the construction of a low-parametric and computationally simple colour model close to the projective one and with better perceptual uniformity, as well as reduced noise heteroscedasticity.

\section{Conclusion}
ProLab is a novel colour coordinate system that is demonstrated to be superior to CIELAB in perceptual uniformity while still preserving colour manifold linearity.
This property is not present in either CIELAB or CAM16-UCS. Reproduction angular errors can be used in proLab, as in linear colour spaces.
ProLab, by design, aligns angular errors of different hues to CIEDE2000 perceptual colour differences unlike the previously mentioned systems.
Further, we demonstrate that image noise in proLab is more homoscedastic than in other standard spaces, including linear ones.
These advantages make proLab a preferred coordinate system in which to perform structural analysis of colour distributions.
Because the incidence of linear manifolds is preserved,  light source direction can be estimated by the intersection of planes formed by the colour distribution of glossy surfaces in proLab. Estimating manifold coordinates with no additional corrections of noise heteroscedasticity should deliver the same or better accuracy when compared to standard linear spaces.
Furthermore, the mutual positions of manifolds, including angles between lines, in proLab are in correspondence with human perception.

The MatLab/Octave implementation is available at \href{https://github.com/konovalenko-iitp/proLab}{https://github.com/konovalenko-iitp/proLab}.

\section*{Acknowledgment}

We would like to thank Prof. Valentina Bozhkova, our colleagues at the 25th Symposium of the International Colour Vision Society, and our colleagues at Huawei Color Constancy \& Multispectral Processing Workshop 2019 for the fruitful discussions of the main idea behind this work.

\balance
\bibliographystyle{IEEEtran}
\bibliography{main}

\begin{IEEEbiography}[{\includegraphics[width=1in,height=1.25in,clip,keepaspectratio]{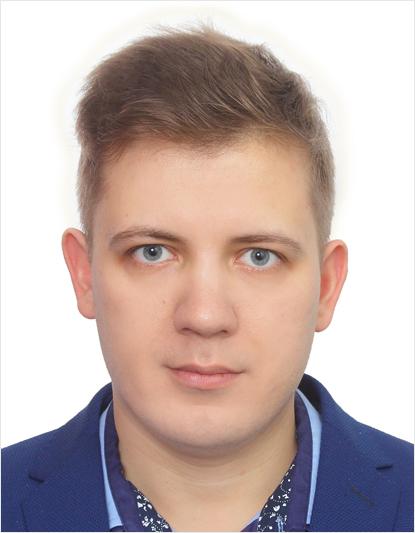}}]
{Ivan~A.~Konovalenko} (ORCID: \href{https://orcid.org/0000-0001-5705-4438}{0000-0001-5705-4438})
was born in Karasuk, Novosibirsk region, USSR in 
1990.
He received the B.S. and M.S. degrees in applied mathematics and physics from the Moscow Institute of Physics and Technology, Dolgoprudny, Russia, in 2014.

From 2009 to 2013, he was a Trainee Researcher with the Predictive Modeling and Optimization Laboratory, Institute for Information Transmission Problems (IITP) of the Russian Academy of Sciences.
Since 2013, he has been an Researcher with the Vision Systems Laboratory, IITP.
He is the author of more than 60 articles.
His research interests include
computer vision,
colorimetry,
projective geometry
and mathematical optimization.
\end{IEEEbiography}

\begin{IEEEbiography}[{\includegraphics[width=1in,height=1.25in,clip,keepaspectratio]{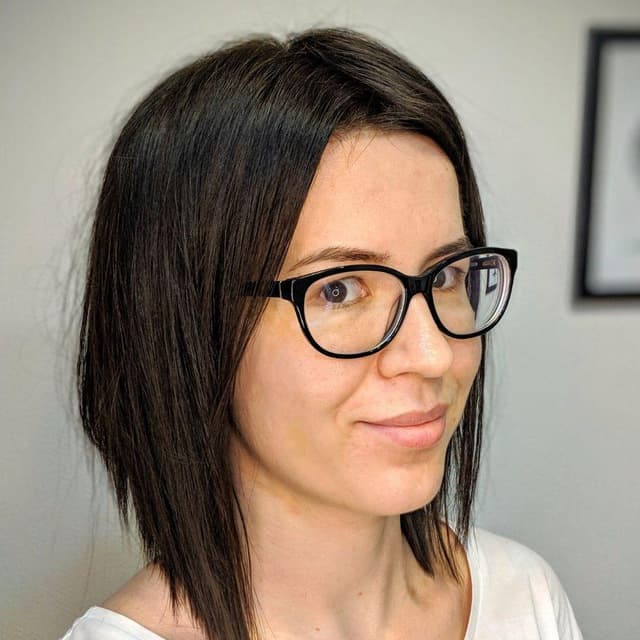}}]
{Anna~A.~Smagina} (ORCID: \href{https://orcid.org/0000-0002-0135-9280}{0000-0002-0135-9280})
received the B.S. degree in applied math and physics from Samara State Aerospace Univeristy, Russia, in 2013, and
the M.S. degree in high energy physics from Moscow Institute of Physics and Technology, Russia in 2016.
From 2017, she has been working as researcher in the Vision Systems Laboratory, Institute for Information Transmission Problems, Russian Academy of Sciences (Kharkevich Institute), Moscow.
Author of more than 30 articles. 
Research interests include technical colour vision and 3D computer vision.
\end{IEEEbiography}

\begin{IEEEbiography}[{\includegraphics[width=1in,height=1.25in,clip,keepaspectratio]{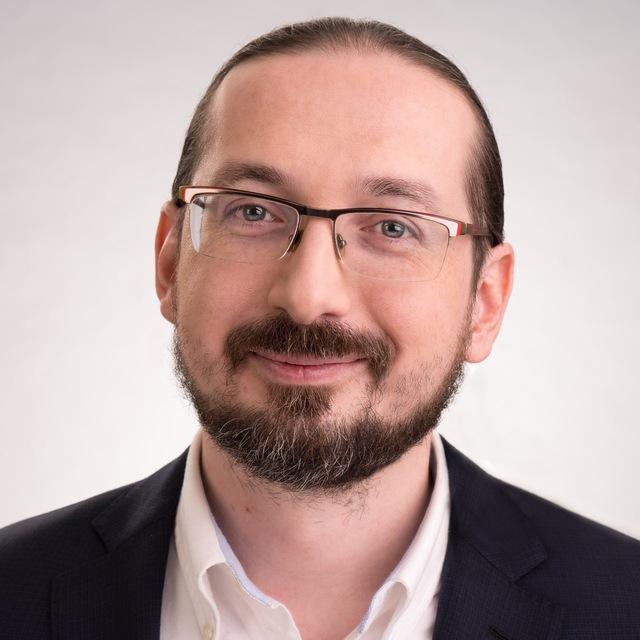}}]
{Dmitry~P.~Nikolaev} (ORCID: \href{https://orcid.org/0000-0001-5560-7668}{0000-0001-5560-7668}) (Member, IEEE)
was born in Moscow, Russia, in 1978. He received the master's degree in physics and the Ph.D. degree in computer science from Moscow State University, Moscow, Russia, in 2000 and 2004, respectively.
Since 2007, he has been the Head of the Vision Systems Laboratory, Institute for Information Transmission Problems, Russian Academy of Sciences (Kharkevich Institute), Moscow, and he has been the CTO of Smart Engines Service LLC, Moscow, since 2016.

Since 2016, he has been an Associate Professor with the Moscow Institute of Physics and Technology (State University), Moscow, teaching the Image Processing and Analysis Course.
He has authored over 250 scientic publications and 10 patents. His research activities are in the area of computer vision with primary application to color image understanding.
Dr. Nikolaev led a team of authors to win the Document Image Binarization Competition (DIBCO), in 2017.
\end{IEEEbiography}

\begin{IEEEbiography}[{\includegraphics[width=1in,height=1.25in,clip,keepaspectratio]{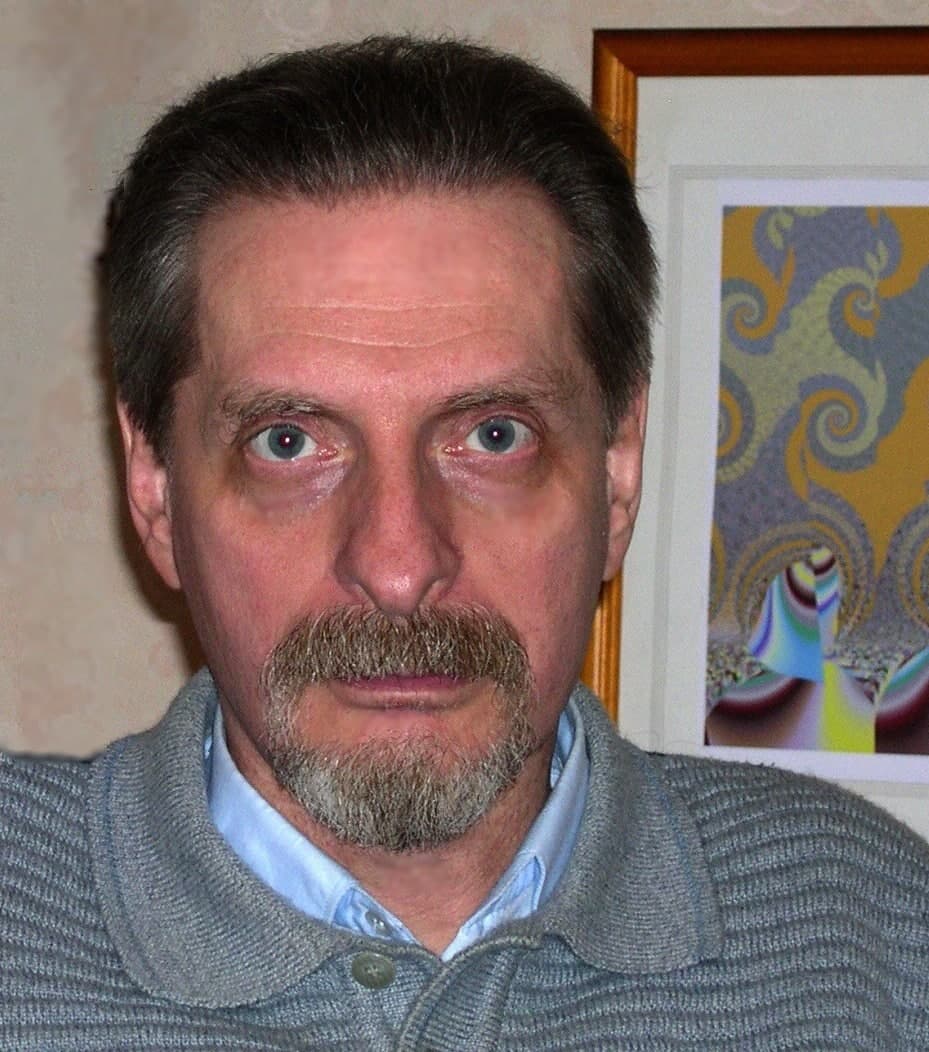}}]
{Petr~P.~Nikolaev} (ORCID: \href{https://orcid.org/0000-0003-3016-3903}{0000-0003-3016-3903}) 
has a degree of Full Doctor of Phys. and Math. Sciences, awarded in 1993 in field of biophysics. He was graduated in Moscow State University in 1966 majoring in physics and got a Ph.D. degree in 1975. At present he has a tenured post of Head Researcher of Vision Systems Laboratory, Institute for Information Transmission Problems, Russian Academy of Sciences (Kharkevich Institute), Moscow and works on the theoretical problems of color and space perception, recognition and 3D representation. His major scientific achievements are related to the fundamental aspects of psychophysics of human vision, pattern recognition and image analysis. Since 2015, he has been an Professor with the Moscow Institute of Physics and Technology (State University), Moscow, teaching the Computer Vision Course.
He has authored over 100 scientic publications and 3 books.
Prof. Nikolaev is a Laureate of the national award “Russia’s outstanding scientist”.
\end{IEEEbiography}

\EOD

\end{document}